\pdfoutput=1

\documentclass[11pt]{article}
\usepackage[dvipsnames]{xcolor}

\usepackage[final]{acl}

\usepackage{times}
\usepackage{latexsym}
\usepackage{graphicx}
\usepackage{booktabs}

\usepackage[T1]{fontenc}

\usepackage[utf8]{inputenc}

\usepackage{microtype}

\usepackage{inconsolata}

\usepackage{array}
\usepackage{amsmath}
\usepackage{amsfonts}
\usepackage{bbm}
\usepackage{listings}
\usepackage{float}
\usepackage{tikz}
\usetikzlibrary{positioning, shapes, calc, fit, decorations.pathmorphing, decorations.shapes, patterns.meta, arrows.meta}
\usepackage{multirow}

\usepackage{pgf} 
\usepackage{colortbl}

\newcommand{\gradientcell}[7]{
    \ifdimcomp{#1pt}{>}{#3 pt}{\cellcolor{#5!100.0!#4!#6}\hspace*{-7.2pt}#7 #1\hspace*{-7.2pt}}{
    \ifdimcomp{#1pt}{<}{#2 pt}{\cellcolor{#5!0.0!#4!#6}\hspace*{-7.2pt}#7 #1\hspace*{-7.2pt}}{
         \pgfmathparse{int(round(100*(#1/(#3-#2))-(#2 *(100/(#3-#2)))))}
        \xdef\tempa{\pgfmathresult}
        \cellcolor{#5!\tempa!#4!#6}\hspace*{-7.2pt}#7 #1\hspace*{-7.2pt}
    }}
 }

\newfloat{lstfloat}{htbp}{lop}
\floatname{lstfloat}{Algorithm}

\lstdefinestyle{mystyle}{   
    keywordstyle=\ttfamily\bfseries,
    numberstyle=\tiny\color{gray},    
    basicstyle=\ttfamily\footnotesize,
    breakatwhitespace=false,         
    breaklines=true,                 
    captionpos=b,                    
    keepspaces=true,                 
    numbers=left,                    
    numbersep=5pt,                  
    showspaces=false,                
    showstringspaces=false,
    showtabs=false,                  
    tabsize=2,
    columns=flexible,
    commentstyle=\footnotesize\color{gray},
    xleftmargin=0.4cm,
}

\let\origfootnote\footnote
\renewcommand{\footnote}[1]{\kern.1em\origfootnote{#1}}
\newcommand{\punctfootnote}[1]{\kern-.1em\origfootnote{#1}}

\newcommand{\ar}[1][3pt]{\mathrel{%
   \vcenter{\hbox{\rule[-.5\fontdimen8\textfont3]{#1}{\fontdimen8\textfont3}}}%
   \mkern-4mu\hbox{\usefont{U}{lasy}{m}{n}\symbol{41}}}}

\DeclareMathOperator{\softmax}{softmax}
\DeclareMathOperator{\norm}{norm}

\usepackage{todonotes}

\title{Lexically Grounded Subword Segmentation}

\author{Jindřich Libovický \and Jindřich Helcl \\
  Charles University, Faculty of Mathematics and Physics \\
  Institute of Formal and Applied Linguistics \\
  V Holešovičkách 2, 180 00 Prague, Czech Republic \\  
  \texttt{\{libovicky, helcl\}@ufal.mff.cuni.cz}}

\begin{document}
\maketitle
\begin{abstract}
We present three innovations in tokenization and subword segmentation. 
First, we propose to use unsupervised morphological analysis with Morfessor as pre-tokenization. 
Second, we present an algebraic method for obtaining subword embeddings grounded in a word embedding space.
Based on that, we design a novel subword segmentation algorithm that uses the embeddings, ensuring that the procedure considers lexical meaning.
Third, we introduce an efficient segmentation algorithm based on a subword bigram model that can be initialized with the lexically aware segmentation method to avoid using Morfessor and large embedding tables at inference time.
We evaluate the proposed approaches using two intrinsic metrics and measure their performance on two downstream tasks: part-of-speech tagging and machine translation.
Our experiments show significant improvements in the morphological plausibility of the segmentation when evaluated using segmentation precision on morpheme boundaries and improved Rényi efficiency in 8 languages.
Although the proposed tokenization methods do not have a large impact on automatic translation quality, we observe consistent performance gains in the arguably more morphological task of part-of-speech tagging.
\end{abstract}

\section{Introduction}

Statistical approaches to subword segmentation are the state of the art in most natural language processing (NLP) applications of neural networks, most notably the Transformer model \citep{vaswani-etal-2017-attention}. 
The Unigram model from SentencePiece \citep{kudo-richardson-2018-sentencepiece} and Byte-Pair Encoding (BPE; \citealp{sennrich-etal-2016-neural}) are among the two most widely employed tokenization techniques.
These methods gained popularity because of their versatility -- they are language-independent and have convenient properties for model training, reducing the vocabulary size while assuring even learning of the token representations.

Despite the indisputable advantages, one aspect of the statistical word segmentation algorithms has remained a thorn in the eyes of many linguistically-oriented researchers: \emph{Subwords do not reflect morphology}.
This problem is especially pronounced in multilingual models, which share a common vocabulary across all languages. Without a careful and balanced data selection, lower-resourced languages tend to have fewer allocated subwords, resulting in a large token-to-word ratio \citep{haddow-etal-2022-survey,limisiewicz-etal-2023-tokenization}.

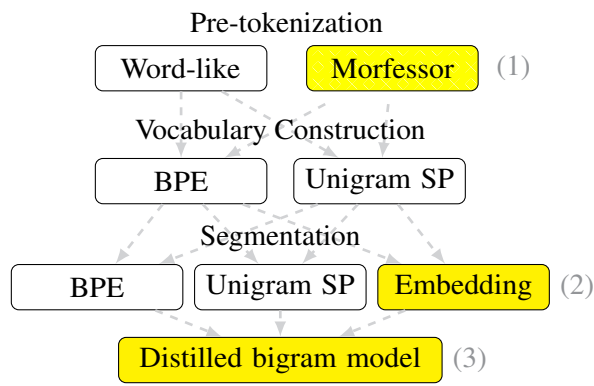
\begin{figure}[t]
\centering\begin{tikzpicture}[
    stepBox/.style={draw, text width=20mm, align=center, minimum height=6mm, rounded corners=1mm},
    connectline/.style={arrows={-latex},color=Gray!40,line width=1pt, dashed},
    numbered/.style={text=Gray}
]

\node (origin) {};

\begin{scope}[local bounding box=pretokenization]
    \node[stepBox, left=1pt of origin] (wordlike) {Word-like};
    \node[stepBox, right=1pt of origin, pattern={Lines[line width=10pt,angle=45]}, pattern color=Yellow] (morfessor) {Morfessor};
\end{scope}

\begin{scope}[local bounding box=vocabLearn]
    \node[stepBox, below left=30pt and 1pt of origin] (bpeVocab) {BPE};
    \node[stepBox, below right=30pt and 1pt of origin] (spVocab) {Unigram SP};
\end{scope}

\begin{scope}[local bounding box=inference]
    \node[stepBox, below =70pt of origin.south] (spInf) {Unigram SP};
    \node[stepBox, left =4pt of spInf] (bpeInf) {BPE};
    \node[stepBox, right =4pt of spInf, fill=Yellow] (embInf) {Embedding};
\end{scope}

\node[stepBox, text width=40mm, below=10pt of spInf, fill=Yellow] (distill) {Distilled bigram model};

\draw[connectline] (wordlike) -- (bpeVocab);
\draw[connectline] (wordlike) -- (spVocab);
\draw[connectline] (morfessor) -- (bpeVocab);
\draw[connectline] (morfessor) -- (spVocab);

\draw[connectline] (bpeVocab) -- (bpeInf);
\draw[connectline] (spVocab) -- (bpeInf);
\draw[connectline] (bpeVocab) -- (spInf);
\draw[connectline] (spVocab) -- (spInf);
\draw[connectline] (bpeVocab) -- (embInf);
\draw[connectline] (spVocab) -- (embInf);

\draw[connectline] (bpeInf) -- (distill);
\draw[connectline] (spInf) -- (distill);
\draw[connectline] (embInf) -- (distill);

\node [above=1pt of pretokenization] {Pre-tokenization};
\node [above=1pt of vocabLearn] {Vocabulary Construction};
\node [above=1pt of inference] {Segmentation};
\node [above=1pt of distill] {};

\node [numbered, right=-2pt of morfessor] {(1)};
\node [numbered, right=-0pt of embInf] {(2)};
\node [numbered, right=-0pt of distill] {(3)};

\end{tikzpicture}
\caption{We organize subword tokenization learning into four steps: pre-tokenization, vocabulary learning, inference, and distillation for efficiency. Steps (1)--(3) highlighted in \colorbox{Yellow}{yellow} are specific contributions of this paper.}\label{fig:scheme}
\end{figure}

We posit that a strong segmentation retains the property of the statistical approaches, i.e., that frequent words are split into fewer tokens than rare words.
However, once a word is split into more tokens, the subword boundaries should ideally match the actual morpheme boundaries.\punctfootnote{We use the word \emph{morpheme} for morphologically motivated subword units. Some theories \citep{zabokrtsky-etal-2022-towards} distinguish \emph{morphs} as surface realizations of abstract morphemes as the smallest units of meaning. Where appropriate, we follow this distinction for clarity. By \emph{morpheme boundaries}, we mean boundaries between morphs within a word.}
We hypothesize that the standard algorithms lack morphology awareness because they do not work with lexical meaning, which is a crucial concept in language morphology.

Following \citet{schmidt2024tokenizaion}, we conceptualize tokenization as a process with three steps (as illustrated in Figure~\ref{fig:scheme}): pre-tokenization, vocabulary construction, and segmentation. Within this conceptual framework,
we propose three innovations throughout the whole process:
\begin{itemize}
    \item[(1)] We consider unsupervised morphological segmentation as an alternative for pre-tokenization.
    \item[(2)] Propose a novel lexically grounded segmentation algorithm based on word and subword embeddings.
    \item[(3)] We propose an efficient statistical segmentation algorithm using subword bigram statistics that can be used to distill complex tokenization pipelines into an efficient algorithm.
\end{itemize}

In Section~\ref{sec:pretok}, we discuss pre-tokenization and vocabulary construction. Besides the standard pre-tokenization, which splits the text into word-like units (words, punctuation, etc.), we also experiment with Morfessor \citep{smit-etal-2014-morfessor}, which we apply on top of the word-like pre-tokenized text.

For lexically grounded segmentation, we derive \emph{a formula for computing subword
embeddings} using a pre-trained word embedding model and a training corpus (Section~\ref{subsec:embeddings}). 
Next, we use the subword embeddings to design \emph{a subword segmentation algorithm based on semantic similarity}
between the word and its subwords (Section~\ref{subsec:segment}). 

Finally, we propose \emph{a subword-bigram-based statistical segmentation algorithm}
that retains the properties of the embedding-based segmentation (Section~\ref{sec:bigram}).
With the bigram-based algorithm, we can have a model for subword segmentation that does not require running Morfessor or storing a large embedding table.

We test our approach using two intrinsic evaluation metrics and two downstream tasks (Section~\ref{sec:evaluation}).
In the intrinsic evaluation, we test our approach on the SIGMORPHON 2018 shared task dataset \citep{batsuren-etal-2022-sigmorphon} and observe significantly better morphological generalization in both proposed algorithms with a fixed vocabulary size. We also measure the Rényi efficiency \citep{renyi1961measures} of the unigram distribution of the segmented text, which has been shown to correlate with downstream model performance \citep{zouhar-etal-2023-tokenization}.
Additionally, we evaluate our segmentation algorithm on Part-of-Speech (POS) Tagging using Universal Dependencies \citep{universal-dependencies}, showing an improvement compared to other segmentations.
Finally, we evaluate our tokenization on machine translation using a simulated low-resource IWSLT 2017 dataset \citep{cettolo-etal-2017-overview} where we reach results comparable with currently used subword tokenizers.

We show the code examples in Appendix~\ref{app:code} and we release the code for the segmentation tool, \textsc{Legros},\punctfootnote{\url{https://github.com/ufal/legros}} as well as the experimental code.\punctfootnote{\url{https://github.com/ufal/legros-paper}}

\section{Pre-tokenization and Vocabulary Construction}%
\label{sec:pretok}

Neural networks can only have limited vocabularies in order $10^4$--$10^5$, which rules out using word-based vocabularies.
A common solution is statistical heuristics that keep frequent words intact and split rare words into smaller units, ensuring that there are no rare tokens,
such that embeddings of all tokens get updated reasonably often.
The most popular methods are Byte-Pair Encoding (BPE; \citealp{sennrich-etal-2016-neural}) based on greedily merging the most frequent token pairs and the Unigram model (as implemented in SentencePiece; \citealp{kudo-2018-subword}) that returns high-probability segmentations using a unigram language model. 
However, these methods manifest low morphological generalization, which in turn might lead to reduced interpretability, compositional generalization, and cross-lingual transfer capabilities.

Perhaps the most straightforward approach for lexically grounded word segmentation is to
use unsupervised morphological analyzers, such as Morfessor. However, direct use of these linguistically motivated tools leads to worse results \citep{machacek2018morphological} and is only beneficial in low-resource scenarios \citep{soulos-etal-2021-structural,gaser-etal-2023-exploring}. Furthermore, morphological analysis does not fully address the problems of rare tokens and vocabulary size.
To address these issues, we propose only using morphological analyzers during pre-tokenization (Step 1 in Figure~\ref{fig:scheme}). After pre-tokenization, we apply the well-established statistical methods for vocabulary construction. This combination ensures that there will be a low number of rare tokens and efficient control of vocabulary size while still preserving the lexical meaning of the subwords. 

\section{Segmentation with Subword Embeddings}\label{sec:segmentation}

In this section, we describe a novel lexically-grounded segmentation method (Step 2 in Figure~\ref{fig:scheme}).

When considering language morphology, we assume the word can be decomposed into several smaller meaningful units that carry the meaning of the original word when combined together. We consider the segmentation of a word to be lexically grounded when it respects the word's meaning and does not introduce subword boundaries in the middle of meaningful units. To find such a segmentation, we need to model the meaning of both words and subword units jointly.\punctfootnote{Linguistic theories often work with the concept of morphs and morphemes as the smallest meaningful units. However, our solution tries to be theory-agnostic, so it can work with any subword units regardless of their theoretical justification.}

A widely used proxy for capturing the lexical meaning of words is word embeddings. To capture the meaning of subwords, we introduce a method to compute subword embeddings in a shared space with the word embeddings (§ \ref{subsec:embeddings}). We also describe a segmentation algorithm that takes the subword embeddings into account (\S~\ref{subsec:segment}).

\subsection{Subword Embeddings}\label{subsec:embeddings}

We obtain the joint embedding model of words and subwords by extending the skip-gram model \citep{mikolov2013w2v} to subword units.
Specifically, we derive a formula for computing the embedding of any substring in a training dataset, situating its representation within the skip-gram model embedding space.

Skip-gram models are trained to produce a probability distribution of words that are likely to appear within a certain context window around a given input word $x$.
When we extend this model to handle substrings, each substring is used to predict the whole words that appear within the context window of any word that contains the substring.
As a result, the embeddings of the substrings are determined by the contexts of the words they are part of.

To compute the subword embeddings, we require a tokenized training dataset $\mathcal{D}$ and a trained skip-gram word embedding model with a vocabulary $\mathcal{V}$. In addition to its input embedding matrix $E \in \mathbb{R}^{|\mathcal{V}|\times d}$ where $d$ is the dimension of the word embedding vectors, we also need the output matrix $W \in \mathbb{R}^{d\times |\mathcal{V}|}$.

\paragraph{The statistics of skip-gram models.}

Using data $\mathcal{D}$, we denote the symmetric word cooccurrence matrix $C\in \mathbb{R}^{|\mathcal{V}|\times|\mathcal{V}|}$ that for each pair of words $x, y \in \mathcal{V}$, $C_{x,y}$ contains the frequency of $x$ and $y$ appearing within the same context window in $\mathcal{D}$.
Then, our method relies on the following observation:
\begin{equation}
\softmax(E W) \approx \norm(C)
\end{equation}
where $\norm$ means row-wise normalization.

This follows from the fact that the skip-gram model optimizes cross-entropy between the predicted distribution of neighboring words and the empirical distribution in the training data. It is usually approximated by stochastic minibatch training with negative sampling instead of computing the full softmax. The empirical distribution can be obtained by normalizing the count matrix $C$, which leads to the following optimization problem: 
\begin{equation}
 \min_{E,W} \operatorname{XENT}(\softmax(EW), \norm(C))   
\end{equation}
By Gibbs inequality, the cross-entropy is minimum if $\softmax(EX) = \norm(C)$. This leads to Equation 1. We use the approximation sign ($\approx$) to stress that stochastic optimization solves the problem only approximately. When training word embeddings, we must find both $E$ and $W$. When extending the model for subwords, we keep the $W$ fixed, and we only need to find the (newly added) subword portion of $E$, which we call $E_s$.

\paragraph{Extension to subwords.}

Next, we choose a set of subwords $\mathcal{S}$. We either select the set of all substrings present in $\mathcal{D}$ up to a certain length, or we use the set of subwords from an existing segmentation.
We then define a segmentation matrix $A \in \mathbb{R}^{|\mathcal{S}|\times |\mathcal{V}|}$ such that:
\begin{equation} \label{eq:a}
A_{s,x} = \begin{cases}
    1, & \text{if $s$ belongs to $x$},\\
    0, & \text{otherwise}.
  \end{cases}
\end{equation} 

Then, the multiplication $AC$ corresponds to the subword-word cooccurrence matrix. 
Thus, we can find the substring embedding matrix $E_s \in \mathbb{R}^{|\mathcal{S}| \times d}$
by solving the following formula:
\begin{equation}
\softmax(E_s W) \approx \norm(AC), 
\end{equation}
which can be solved using a least-square approximation as:
\begin{equation}
E_s = \log(\norm(AC)) W_{\mathrm{right}}^{-1} \label{eq:subwemb}
\end{equation}
where $W_{\mathrm{right}}^{-1}$ is the right-inverse of the skip-gram's output matrix $W$.

\subsection{Segmentation}\label{subsec:segment}

In this section, we apply the subword embedding model to lexically grounded subword segmentation.
We propose an algorithm based on the word-subword similarities within the shared embedding space.
Following the Unigram model from SentencePiece \citep{kudo-2018-subword}, which searches for a segmentation that maximizes the probability under a subword unigram model, we use a dynamic programming algorithm (shown in Algorithm~\ref{alg:embedding} in Appendix~\ref{app:code})
to find the segmentation (sequence of subwords) that maximizes a similarity-based score.

Formally, for a word $x$ and a segmentation $s_1, s_2, \ldots, s_n$, the similarity 
score is the sum of cosine similarities between the embedding of $x$ and the embeddings of each of the subwords $s_i$, minus a length penalty of $\alpha$ per each subword:
\begin{equation}
    \sum_{i=1}^{n} \frac{(E(x)) \cdot (E_s (s_i))}{\|E(x)\|\cdot\|E_s(s_i)\|} - \alpha.
\end{equation}
Increasing the value of $\alpha$ forces the algorithm to use fewer subwords.
In other words, $\alpha$ controls what weight we put to the semantic similarity and what weight we put to minimize the number of subwords.
Based on preliminary results, we set $\alpha$ to 1 and keep it fixed in all experiments.
 
Unlike the Unigram segmentation, the subword scores are not static but depend on the segmented word. Therefore, the segmentation can be viewed as a word-specific unigram model.

As stated in the previous section, the computation of the subword embeddings requires an existing subword vocabulary $\mathcal{S}$ and the segmentation matrix $A$. 
We initialize $\mathcal{S}$ with the set of subwords used by another segmentation algorithm. We only set $A_{s,x} = 1$ when $s$ has been used as a subword of $x$.

After initialization, we iteratively refine the segmentation in two alternating steps until convergence.
\begin{enumerate}

\item For a segmentation matrix $A$, calculate subword embeddings $E_s$ (Equation~\ref{eq:subwemb}).

\item For subword embeddings $E_s$, find a new best segmentation and update the segmentation matrix $A$ accordingly. Note that subwords not used in this step are never used again, and therefore, the vocabulary shrinks as the algorithm proceeds.

\end{enumerate}

\section{Bigram model}\label{sec:bigram}

The segmentation algorithm described in the previous section has several drawbacks: It requires storing relatively large embedding tables for words and subwords and does not generalize for OOV words without embeddings. Moreover, pre-tokenization with Morfessor requires running language-specific models, making the segmentation more computationally demanding than the established method.

We avoid this drawback by introducing an alternative segmentation algorithm based on subword bigram statistics. It is a straightforward generalization of the commonly used Unigram model. At inference time, we search for a segmentation that maximizes probability predicted by a subword bigram model instead of a unigram model. The optimization problem is solvable using dynamic programming, similar to the Unigram model. However, the algorithm has a quadratic complexity in the segmented string length. Therefore, we propose using a linear-time beam search algorithm that only considers $k$ best segmentations in each step. The full algorithm is described in Algorithm~\ref{alg:bigram} in Appendix~\ref{app:code}.

We use the subword bigram statistic obtained by counting subword bigram and unigram frequencies in a corpus tokenized by a tokenizer that we want to distill into the bigram model.
To account for unknown bigrams encountered during inference, we need to eliminate zero probabilities from the bigram distribution. To this end, we apply Laplacian smoothing, i.e., we increase the frequency of every bigram $(s_i|s_{i-1})$ by one. Additionally, if $s_{i-1}$ is an unknown unigram, we assign the unigram probability of $s_i$ to the bigram. If both $s_i$ and $s_{i-1}$ are unknown unigrams, we assign uniform probability $1/|\mathcal{S}|$ to the bigram.

\section{Experiments}%
\label{sec:experiments}

We evaluate our proposed methods intrinsically using morpheme boundary precision and Rényi efficiency, as well as extrinsically on two downstream tasks: part-of-speech tagging and machine translation.

\subsection{Intrinsic Evaluation}%
\label{sec:evaluation}

We evaluate the capability of our framework to capture morphological boundaries and compare it with commonly used segmentation methods. Our main evaluation metrics are precision on morpheme boundaries (given a fixed vocabulary size budget) and Rényi efficiency \citep{renyi1961measures} of the token distribution, which was shown to be a good predictor of downstream performance of a tokenizer \citep{zouhar-etal-2023-tokenization}.

\paragraph{Test data.} For the morpheme boundary evaluation, we use the test set from the SIGMORPHON 2022 Shared Task on Morpheme Segmentation \citep{batsuren-etal-2022-sigmorphon}, which contains test data for nine languages (Czech, English, Spanish, Hungarian, French, Italian, Russian, Latin, Mongolian). We omit Latin due to the lack of resources for training word embeddings. 
Except for Czech (which contains surface-level segmentation into morphs), each test set consists of word decompositions into morphemes. This means that the original words cannot be reconstructed by simply concatenating the morphemes.
To be able to evaluate word segmentations in all languages, we use a set of heuristic rules to map the morphemes to the surface form. 

To measure the Rényi efficiency of the token distribution, we use 4,000 sentences randomly sampled from the (plain text) training data described in the following paragraph.

\paragraph{Experimental settings.} We use the skip-gram model from FastText \citep{bojanowski-etal-2017-enriching} to train the word embeddings. For all languages except Mongolian, we train the model on 50M sentences from NewsCrawl \citep{kocmi-etal-2022-findings}. We use 15M sentences from CC-100 \citep{conneau-etal-2020-unsupervised} for Mongolian. We lowercase and pre-tokenize the text using Sacremoses,\punctfootnote{\url{https://github.com/hplt-project/sacremoses}} and for experiments with Morfessor pre-tokenization, we train Morfessor \citep{smit-etal-2014-morfessor} with the default parameters. We apply Morfessor on already pre-tokenized text as a second step. We use a vocabulary size of 200k, an embedding dimension of 200, and a window size of 5. We train the embeddings for 10 epochs for both pre-tokenization setups.

As a baseline, we prepare BPE and Unigram tokenizers with vocabularies 1k, 2k, 4k, 8k, 16k, 24k, 32k, and 48k using the same plain text dataset.

We use the segmentation from the BPE and Unigram subwords to initialize the matrix $A$ from Equation \ref{eq:a} and iterate our algorithm. Finally, we use the bigram statistics from 200k embedding vocabulary and segment the test set using the subword bigram language model.

\paragraph{Segmentation evaluation.} Unlike the original SIGMORPHON shared task evaluation, where the evaluation metric was the F$_1$ score measured on the morphemes themselves, we measure the morpheme boundary precision for a given vocabulary size. We believe this setup best captures the use of subword tokenizers in neural networks where we have a vocabulary budget given by the model architecture. However, we do report also recall and F$_1$ score for completeness.

\paragraph{Results.}

\begin{table*}[ht]

\footnotesize\centering
\newcommand{\PrecisioncsThree}[2]{\gradientcell{#1}{76.5}{91.0}{Cyan}{Yellow}{60}{#2}}
\newcommand{\RenyicsThree}[2]{\gradientcell{#1}{0.419}{0.458}{Cyan}{Yellow}{60}{#2}}
\newcommand{\PrecisionenThree}[2]{\gradientcell{#1}{56.6}{72.0}{Cyan}{Yellow}{60}{#2}}
\newcommand{\RenyienThree}[2]{\gradientcell{#1}{0.429}{0.442}{Cyan}{Yellow}{60}{#2}}
\newcommand{\PrecisionesThree}[2]{\gradientcell{#1}{60.6}{66.4}{Cyan}{Yellow}{60}{#2}}
\newcommand{\RenyiesThree}[2]{\gradientcell{#1}{0.396}{0.429}{Cyan}{Yellow}{60}{#2}}
\newcommand{\PrecisionfrThree}[2]{\gradientcell{#1}{57.1}{67.5}{Cyan}{Yellow}{60}{#2}}
\newcommand{\RenyifrThree}[2]{\gradientcell{#1}{0.421}{0.454}{Cyan}{Yellow}{60}{#2}}
\newcommand{\PrecisionhuThree}[2]{\gradientcell{#1}{77.0}{85.9}{Cyan}{Yellow}{60}{#2}}
\newcommand{\RenyihuThree}[2]{\gradientcell{#1}{0.373}{0.403}{Cyan}{Yellow}{60}{#2}}
\newcommand{\PrecisionitThree}[2]{\gradientcell{#1}{52.9}{63.2}{Cyan}{Yellow}{60}{#2}}
\newcommand{\RenyiitThree}[2]{\gradientcell{#1}{0.437}{0.46}{Cyan}{Yellow}{60}{#2}}
\newcommand{\PrecisionmnThree}[2]{\gradientcell{#1}{78.8}{92.5}{Cyan}{Yellow}{60}{#2}}
\newcommand{\RenyimnThree}[2]{\gradientcell{#1}{0.468}{0.503}{Cyan}{Yellow}{60}{#2}}
\newcommand{\PrecisionruThree}[2]{\gradientcell{#1}{61.7}{73.6}{Cyan}{Yellow}{60}{#2}}
\newcommand{\RenyiruThree}[2]{\gradientcell{#1}{0.414}{0.461}{Cyan}{Yellow}{60}{#2}}

\begin{tabular}{lll cccccccc@{\hskip 18pt}cccccccc}
\toprule

\multicolumn{2}{l}{Vocab.} & Inf. & \multicolumn{8}{c}{Morpheme boundary precision} & \multicolumn{8}{c}{Rényi efficiency} \\ \cmidrule(lr{18pt}){4-11} \cmidrule(lr){12-19}
& & & cs & en & es & fr & hu & it & mn & ru &  cs & en & es & fr & hu & it & mn & ru \\ \midrule
\multirow{6}{*}{\rotatebox{90}{Word-like}} & \multirow{3}{*}{BPE} & Orig & \PrecisioncsThree{76.5}{} & \PrecisionenThree{56.6}{} & \PrecisionesThree{60.6}{} & \PrecisionfrThree{57.1}{} & \PrecisionhuThree{77.0}{} & \PrecisionitThree{52.9}{} & \PrecisionmnThree{78.8}{} & \PrecisionruThree{61.7}{} & \RenyicsThree{.419}{} & \RenyienThree{.429}{} & \RenyiesThree{.396}{} & \RenyifrThree{.421}{} & \RenyihuThree{.373}{} & \RenyiitThree{.437}{} & \RenyimnThree{.470}{} & \RenyiruThree{.414}{}\\
& & Emb. & \PrecisioncsThree{78.9}{} & \PrecisionenThree{65.8}{} & \PrecisionesThree{63.9}{} & \PrecisionfrThree{63.3}{} & \PrecisionhuThree{82.4}{} & \PrecisionitThree{58.3}{} & \PrecisionmnThree{88.9}{} & \PrecisionruThree{64.2}{} & \RenyicsThree{.422}{} & \RenyienThree{.435}{} & \RenyiesThree{.403}{} & \RenyifrThree{.427}{} & \RenyihuThree{.387}{} & \RenyiitThree{.443}{} & \RenyimnThree{.479}{} & \RenyiruThree{.424}{}\\
& & Big. & \PrecisioncsThree{79.4}{} & \PrecisionenThree{66.1}{} & \PrecisionesThree{63.2}{} & \PrecisionfrThree{62.9}{} & \PrecisionhuThree{81.5}{} & \PrecisionitThree{58.2}{} & \PrecisionmnThree{88.1}{} & \PrecisionruThree{66.1}{} & \RenyicsThree{.423}{} & \RenyienThree{.435}{} & \RenyiesThree{.404}{} & \RenyifrThree{.428}{} & \RenyihuThree{.388}{} & \RenyiitThree{.444}{} & \RenyimnThree{.480}{} & \RenyiruThree{.425}{}\\ \cmidrule(l){2-19}
& \multirow{3}{*}{Uni.} & Orig & \PrecisioncsThree{84.3}{} & \PrecisionenThree{64.6}{} & \PrecisionesThree{63.1}{} & \PrecisionfrThree{64.7}{} & \PrecisionhuThree{80.5}{} & \PrecisionitThree{53.3}{} & \PrecisionmnThree{90.4}{} & \PrecisionruThree{66.8}{} & \RenyicsThree{.424}{} & \RenyienThree{.432}{} & \RenyiesThree{.398}{} & \RenyifrThree{.425}{} & \RenyihuThree{.382}{} & \RenyiitThree{.442}{} & \RenyimnThree{.478}{} & \RenyiruThree{.423}{}\\
& & Emb. & \PrecisioncsThree{87.0}{} & \PrecisionenThree{68.3}{} & \PrecisionesThree{65.2}{} & \PrecisionfrThree{66.5}{} & \PrecisionhuThree{82.6}{} & \PrecisionitThree{57.0}{} & \PrecisionmnThree{89.8}{} & \PrecisionruThree{67.6}{} & \RenyicsThree{.424}{} & \RenyienThree{.437}{} & \RenyiesThree{.407}{} & \RenyifrThree{.433}{} & \RenyihuThree{.390}{} & \RenyiitThree{.447}{} & \RenyimnThree{.468}{} & \RenyiruThree{.431}{}\\
& & Big. & \PrecisioncsThree{86.8}{} & \PrecisionenThree{68.8}{} & \PrecisionesThree{64.4}{} & \PrecisionfrThree{66.2}{} & \PrecisionhuThree{82.2}{} & \PrecisionitThree{57.3}{} & \PrecisionmnThree{89.3}{} & \PrecisionruThree{69.1}{} & \RenyicsThree{.425}{} & \RenyienThree{.437}{} & \RenyiesThree{.408}{} & \RenyifrThree{.434}{} & \RenyihuThree{.391}{} & \RenyiitThree{.448}{} & \RenyimnThree{.469}{} & \RenyiruThree{.433}{}\\ \midrule
\multirow{6}{*}{\rotatebox{90}{Morfessor}} & \multirow{3}{*}{BPE} & Orig & \PrecisioncsThree{88.4}{} & \PrecisionenThree{70.7}{} & \PrecisionesThree{66.4}{\bf } & \PrecisionfrThree{66.4}{} & \PrecisionhuThree{82.3}{} & \PrecisionitThree{63.2}{\bf } & \PrecisionmnThree{90.7}{} & \PrecisionruThree{69.1}{} & \RenyicsThree{.449}{} & \RenyienThree{.437}{} & \RenyiesThree{.422}{} & \RenyifrThree{.446}{} & \RenyihuThree{.391}{} & \RenyiitThree{.455}{} & \RenyimnThree{.497}{} & \RenyiruThree{.451}{}\\
& & Emb. & \PrecisioncsThree{88.9}{} & \PrecisionenThree{72.0}{\bf } & \PrecisionesThree{66.3}{} & \PrecisionfrThree{67.0}{} & \PrecisionhuThree{84.9}{} & \PrecisionitThree{62.0}{} & \PrecisionmnThree{92.5}{\bf } & \PrecisionruThree{71.5}{} & \RenyicsThree{.451}{} & \RenyienThree{.440}{} & \RenyiesThree{.425}{} & \RenyifrThree{.449}{} & \RenyihuThree{.401}{} & \RenyiitThree{.457}{} & \RenyimnThree{.500}{} & \RenyiruThree{.456}{}\\
& & Big. & \PrecisioncsThree{88.7}{} & \PrecisionenThree{69.9}{} & \PrecisionesThree{66.2}{} & \PrecisionfrThree{67.5}{\bf } & \PrecisionhuThree{84.3}{} & \PrecisionitThree{62.8}{} & \PrecisionmnThree{91.8}{} & \PrecisionruThree{71.2}{} & \RenyicsThree{.452}{} & \RenyienThree{.440}{} & \RenyiesThree{.426}{} & \RenyifrThree{.449}{} & \RenyihuThree{.400}{} & \RenyiitThree{.458}{} & \RenyimnThree{.500}{} & \RenyiruThree{.457}{}\\ \cmidrule(l){2-19}
& \multirow{3}{*}{Uni.} & Orig & \PrecisioncsThree{89.4}{} & \PrecisionenThree{70.3}{} & \PrecisionesThree{65.3}{} & \PrecisionfrThree{65.4}{} & \PrecisionhuThree{84.0}{} & \PrecisionitThree{61.4}{} & \PrecisionmnThree{90.1}{} & \PrecisionruThree{70.6}{} & \RenyicsThree{.457}{} & \RenyienThree{.441}{} & \RenyiesThree{.426}{} & \RenyifrThree{.452}{} & \RenyihuThree{.398}{} & \RenyiitThree{.460}{\bf } & \RenyimnThree{.503}{\bf } & \RenyiruThree{.461}{\bf }\\
& & Emb. & \PrecisioncsThree{91.0}{\bf } & \PrecisionenThree{70.3}{} & \PrecisionesThree{65.0}{} & \PrecisionfrThree{65.7}{} & \PrecisionhuThree{85.9}{\bf } & \PrecisionitThree{61.7}{} & \PrecisionmnThree{91.0}{} & \PrecisionruThree{73.6}{\bf } & \RenyicsThree{.457}{} & \RenyienThree{.441}{} & \RenyiesThree{.429}{\bf } & \RenyifrThree{.454}{\bf } & \RenyihuThree{.403}{\bf } & \RenyiitThree{.460}{\bf } & \RenyimnThree{.496}{} & \RenyiruThree{.458}{}\\
& & Big. & \PrecisioncsThree{90.2}{} & \PrecisionenThree{69.7}{} & \PrecisionesThree{65.2}{} & \PrecisionfrThree{66.4}{} & \PrecisionhuThree{85.0}{} & \PrecisionitThree{61.6}{} & \PrecisionmnThree{90.7}{} & \PrecisionruThree{72.3}{} & \RenyicsThree{.458}{\bf } & \RenyienThree{.442}{\bf } & \RenyiesThree{.429}{\bf } & \RenyifrThree{.454}{\bf } & \RenyihuThree{.403}{\bf } & \RenyiitThree{.460}{\bf } & \RenyimnThree{.496}{} & \RenyiruThree{.460}{}\\

\bottomrule
\end{tabular}

\caption{Morpheme boundary precision on the SIGMORPHON 2018 test set and Rényi efficiency estimated on 4k plain text sentences for tokenizers with 32k-sized vocabularies. The best results in each column are bolded. The blue-yellow scale is fit to the value range per column. Results for 24k and 40k vocabularies are in Appendix in Table~\ref{tab:intrinsicAppendix}.}\label{tab:intrinsic}

\end{table*}

\begin{figure*}[ht]

\includegraphics[width=\textwidth]{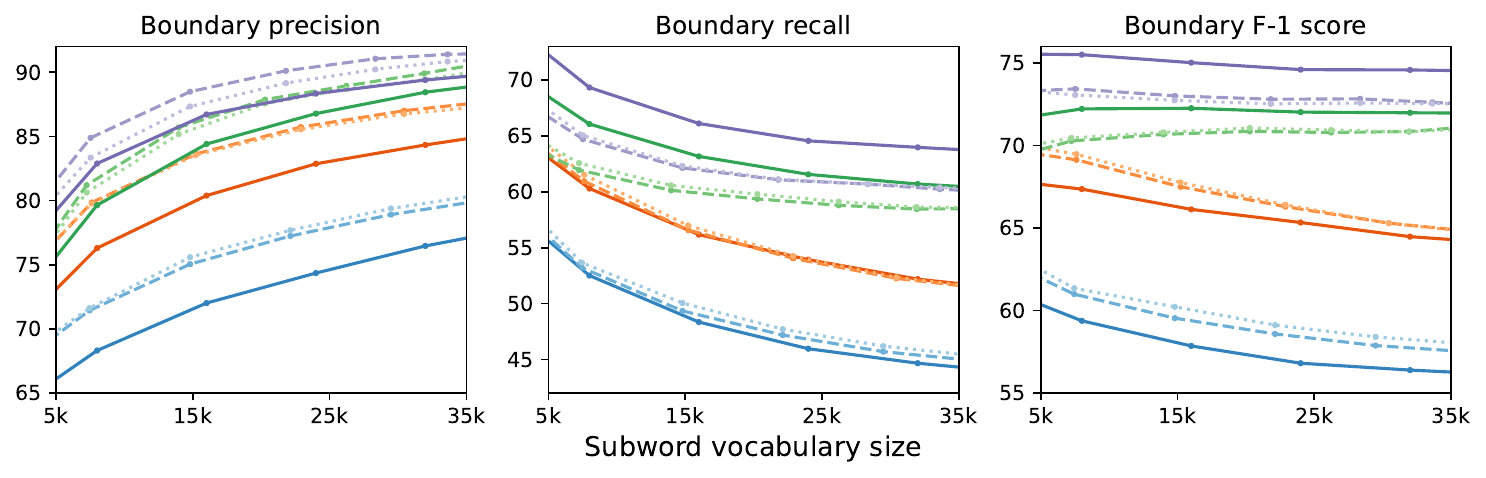}

\definecolor{plot1}{rgb}{0.19215686274509805, 0.5098039215686274, 0.7411764705882353}
\definecolor{plot2}{rgb}{0.9019607843137255, 0.3333333333333333, 0.050980392156862744}
\definecolor{plot3}{rgb}{0.19215686274509805, 0.6392156862745098, 0.32941176470588235}
\definecolor{plot4}{rgb}{0.4588235294117647, 0.4196078431372549, 0.6941176470588235}

\sf\footnotesize\centering
\textcolor{plot1}{$\bullet$} Word + BPE \quad \textcolor{plot2}{$\bullet$} Word + Unigram \quad \textcolor{plot3}{$\bullet$} Morfessor + BPE \quad \textcolor{plot4}{$\bullet$} Morfessor + Unigram \\
\tikz[baseline=-0.5ex]\draw [thick] (0,0) -- (0.5,0); Original \quad \tikz[baseline=-0.5ex]\draw [thick,dashed] (0,0) -- (0.5,0); Embedding-based inference \quad \tikz[baseline=-0.5ex]\draw [thick,dotted] (0,0) -- (.5,0); Bigram model

\caption{Boundary precision, recall, and F$_1$ score for Czech in the SIGMORPHON 2018 test set for different vocabulary sizes. For more other languages, see Figure~\ref{fig:allIntrinsic} in the Appendix.}\label{fig:results}

\end{figure*}

The main results for the 32k vocabulary are presented in Table~\ref{tab:intrinsic}. Across all languages, Unigram reaches better precision than BPE, consistently with previous work \citep{batsuren-etal-2022-sigmorphon}. Pre-tokenization using Morfessor consistently outperforms word-like pre-tokenization across all languages in morpheme boundary precision. Using lexically grounded embedding-based segmentation improves compared to the default BPE and Unigram segmentation algorithms. The difference is more pronounced with the word-like pre-tokenization. Distillation into the bigram model usually leads to a small decrease in the boundary precision. The performance of BPE and the Unigram model for vocabulary construction is language-dependent.

The Rényi efficiency is significantly higher for Morfessor pre-tokenization. Unlike morpheme boundary precision, distilling the embedding-based segmentation into a bigram model has almost no effect on Rényi efficiency. Segmentation based on the Unigram model vocabulary achieves the best results.

Figure~\ref{fig:results} shows morpheme boundary precision, recall, and F$_1$ score for Czech for different vocabulary sizes; additional languages are presented in the Appendix in Figure~\ref{fig:allIntrinsic}. The boundary precision increases with the increasing vocabulary size, whereas the recall has the opposite trend. Our segmentation methods improve the boundary precision in all cases. Word-like pre-tokenization has a negligible effect on recall. On the other hand, adding Morfessor to pre-tokenization decreases recall.

We also show a random sample of segmented Czech, English, and French words in the Appendix in Table \ref{tab:examples}.

\subsection{POS Tagging Evaluation}\label{sec:pos}

\begin{table*}[ht]
\footnotesize\centering
\newcommand{\PosAcccs}[2]{\gradientcell{#1}{96.507}{98.224}{Cyan}{Yellow}{60}{#2}}
\newcommand{\PosAccen}[2]{\gradientcell{#1}{92.551}{93.973}{Cyan}{Yellow}{60}{#2}}
\newcommand{\PosAcces}[2]{\gradientcell{#1}{94.926}{95.72800000000001}{Cyan}{Yellow}{60}{#2}}
\newcommand{\PosAccfr}[2]{\gradientcell{#1}{96.638}{97.33500000000001}{Cyan}{Yellow}{60}{#2}}
\newcommand{\PosAcchu}[2]{\gradientcell{#1}{78.641}{91.638}{Cyan}{Yellow}{60}{#2}}
\newcommand{\PosAccit}[2]{\gradientcell{#1}{96.947}{97.759}{Cyan}{Yellow}{60}{#2}}
\newcommand{\PosAccru}[2]{\gradientcell{#1}{94.658}{97.531}{Cyan}{Yellow}{60}{#2}}
\newcommand{\Aggr}[2]{\gradientcell{#1}{-2.013159192681013}{0.7453248272368196}{Cyan}{Yellow}{60}{#2}}
\newcolumntype{C}{>{\centering}m{15pt}}\begin{tabular}{lll CCCCCCC@{\hskip 18pt}c}
\toprule
\multicolumn{3}{l}{Tokenization} &cs & en & es & fr & hu & it & ru & Aggr.\\ \midrule
\multicolumn{3}{l}{Word vocab} &  \PosAcccs{96.16}{} & \PosAccen{92.07}{} & \PosAcces{94.43}{} & \PosAccfr{96.14}{} & \PosAcchu{79.44}{} & \PosAccit{96.45}{} & \PosAccru{94.16}{} & \Aggr{-2.013}{} \\
\multicolumn{3}{l}{Morfessor} &  \PosAcccs{96.01}{} & \PosAccen{92.05}{} & \PosAcces{94.61}{} & \PosAccfr{96.19}{} & \PosAcchu{78.14}{} & \PosAccit{96.64}{} & \PosAccru{94.48}{} & \Aggr{-1.902}{} \\ \midrule
\multirow{4}{*}{\rotatebox[origin=c]{90}{Word-like}} & \multirow{2}{*}{BPE} & Orig. &  \PosAcccs{98.17}{} & \PosAccen{93.73}{} & \PosAcces{95.50}{} & \PosAccfr{97.16}{} & \PosAcchu{87.76}{} & \PosAccit{97.47}{} & \PosAccru{97.38}{} & \Aggr{0.340}{\hphantom{-}} \\
& & Ours &  \PosAcccs{98.19}{} & \PosAccen{93.78}{} & \PosAcces{95.58}{} & \PosAccfr{97.23}{} & \PosAcchu{88.88}{} & \PosAccit{97.56}{} & \PosAccru{97.40}{} & \Aggr{0.471}{\hphantom{-}} \\ \cmidrule{2-11}
& \multirow{2}{*}{Uni.} & Orig. &  \PosAcccs{98.09}{} & \PosAccen{93.50}{} & \PosAcces{95.41}{} & \PosAccfr{97.00}{} & \PosAcchu{88.57}{} & \PosAccit{97.41}{} & \PosAccru{97.30}{} & \Aggr{0.187}{\hphantom{-}} \\
& & Ours &  \PosAcccs{98.17}{} & \PosAccen{93.76}{} & \PosAcces{95.56}{} & \PosAccfr{97.11}{} & \PosAcchu{89.68}{} & \PosAccit{97.58}{} & \PosAccru{97.43}{} & \Aggr{0.447}{\hphantom{-}} \\ \midrule
\multirow{4}{*}{\rotatebox[origin=c]{90}{Morfessor}} & \multirow{2}{*}{BPE} & Orig. &  \PosAcccs{98.18}{} & \PosAccen{93.91}{} & \PosAcces{95.44}{} & \PosAccfr{97.21}{} & \PosAcchu{90.92}{} & \PosAccit{97.48}{} & \PosAccru{97.39}{} & \Aggr{0.473}{\hphantom{-}} \\
& & Ours &  \PosAcccs{98.21}{\bf } & \PosAccen{93.96}{\bf } & \PosAcces{95.72}{\bf } & \PosAccfr{97.33}{\bf } & \PosAcchu{91.63}{\bf } & \PosAccit{97.74}{} & \PosAccru{97.52}{\bf } & \Aggr{0.745}{\bf \hphantom{-}} \\ \cmidrule{2-11}
& \multirow{2}{*}{Uni.} & Orig. &  \PosAcccs{98.04}{} & \PosAccen{93.86}{} & \PosAcces{95.66}{} & \PosAccfr{97.16}{} & \PosAcchu{91.12}{} & \PosAccit{97.61}{} & \PosAccru{97.35}{} & \Aggr{0.541}{\hphantom{-}} \\
& & Ours &  \PosAcccs{98.11}{} & \PosAccen{93.95}{} & \PosAcces{95.72}{} & \PosAccfr{97.29}{} & \PosAcchu{91.51}{} & \PosAccit{97.75}{\bf } & \PosAccru{97.52}{} & \Aggr{0.712}{\hphantom{-}} \\ \bottomrule
\end{tabular}

\caption{Test accuracies of POS tagging. The final column shows the averaged normalized accuracy (after subtracting the language-specific mean and dividing by the language-specific standard deviation). The blue-yellow scale is fit to the value range per column. More detailed results and additional baselines are in Table~\ref{tab:pos_appendix} in the Appendix. }\label{tab:pos}
\end{table*}

In our first extrinsic evaluation, we experiment with POS tagging as a simple task that directly involves language morphology.

\paragraph{Data.} We use Universal Dependency (UD) Corpora \citep{universal-dependencies} for the languages from the intrinsic evaluation except for Mongolian, which does not have a UD corpus. See Table~\ref{tab:ud_data} in the Appendix for details of the corpora. 

\paragraph{Model details.} We train an LSTM-based tagger. We use an embedding layer of 300, two bidirectional LSTM layers \citep{lstm} of dimension 600, and a final projection into 18 POS tags. We use a batch size of 256 sentences and train for 3,200 steps using the Adam optimizer \citep{kingma2015adam} with a learning rate of 0.01. We select the best weights based on the loss on the development set. We prepend each word with a special word-separator token for subword segmentation and copy the POS tag to all its subwords. At inference time, we predict the tag from a distribution that averages the predictions for the individual subwords.
We are aware that there are methods that would improve the performance of the tagger trained from scratch, e.g., including character-level features and using pre-trained word embeddings. In our experiments, we are mainly interested in how informative the segmentation is for the tagger.

\paragraph{Data preparation.} We experiment with several segmentation methods. As a baseline, we use the word segmentation provided in UD and word segmented using Morfessor. Further, we experimented with word-like pre-tokenization, Morfessor pre-tokenization, and BPE and Unigram for vocabulary construction. For segmentation, we tested both the original subword segmentation corresponding to BPE and the Unigram model (denoted as Orig. in the results) and distilled bigram models created via the lexically grounded embedding-based segmentation (denoted as Ours in the results).

\paragraph{Results.} The results are presented in Table~\ref{tab:pos} (with more details in Table~\ref{tab:pos_appendix} in the Appendix). In general, subword-based segmentation significantly outperforms word-like and Morfessor-based models. Morfessor pre-tokenization is slightly better than word-like pre-tokenization only in all languages, with a particularly pronounced difference in Hungarian, the only language in our test sets with agglutinative morphology. Our segmentation algorithm consistently improves over the default BPE and Unigram algorithms. The overall best tokenization approach combines the Morfessor pre-tokenization followed by the BPE algorithm for vocabulary construction and our bigram-based segmentation. 

\subsection{Machine Translation Evaluation}\label{sec:mt}

\begin{table}

\footnotesize\centering
\newcommand{\Chrfmean}[2]{\gradientcell{#1}{-1.7636846187062625}{1.44629319643834}{Cyan}{Yellow}{60}{#2}}
\begin{tabular}{lll ccc@{\hskip 18pt}c}
\toprule
\multicolumn{3}{l}{\multirow{2}{*}{Tokenization}} & \multicolumn{3}{c}{Vocabulary\hspace*{15pt}} & \multirow{2}{*}{Avg.} \\ \cmidrule(lr{18pt}){4-6}
& & & 4k & 8k & 16k & \\ \midrule

\multirow{4}{*}{\rotatebox[origin=c]{90}{Word-like}} & \multirow{2}{*}{BPE} & Orig. & \Chrfmean{0.0}{\hphantom{-}} & \Chrfmean{0.4}{\hphantom{-}} & \Chrfmean{0.7}{\hphantom{-}} & \Chrfmean{0.4}{\hphantom{-}} \\
& & Ours & \Chrfmean{-0.0}{} & \Chrfmean{0.5}{\hphantom{-}} & \Chrfmean{0.8}{\hphantom{-}} & \Chrfmean{0.4}{\hphantom{-}} \\ \cmidrule{2-7}
& \multirow{2}{*}{Uni.} & Orig. & \Chrfmean{-0.0}{} & \Chrfmean{0.9}{\hphantom{-}} & \Chrfmean{0.9}{\hphantom{-}} & \Chrfmean{0.6}{\hphantom{-}} \\
& & Ours & \Chrfmean{-0.2}{} & \Chrfmean{0.5}{\hphantom{-}} & \Chrfmean{0.6}{\hphantom{-}} & \Chrfmean{0.3}{\hphantom{-}} \\ \midrule
\multirow{4}{*}{\rotatebox[origin=c]{90}{Morfessor}} & \multirow{2}{*}{BPE} & Orig. & \Chrfmean{-1.0}{} & \Chrfmean{-0.8}{} & \Chrfmean{-0.7}{} & \Chrfmean{-0.9}{} \\
& & Ours & \Chrfmean{-0.2}{} & \Chrfmean{0.3}{\hphantom{-}} & \Chrfmean{0.5}{\hphantom{-}} & \Chrfmean{0.2}{\hphantom{-}} \\ \cmidrule{2-7}
& \multirow{2}{*}{Uni.} & Orig. & \Chrfmean{-1.3}{} & \Chrfmean{-0.9}{} & \Chrfmean{-0.9}{} & \Chrfmean{-1.0}{} \\
& & Ours & \Chrfmean{-0.1}{} & \Chrfmean{0.3}{\hphantom{-}} & \Chrfmean{-0.2}{} & \Chrfmean{-0.0}{} \\ \bottomrule
\end{tabular}

\caption{Mean deviation from the average chrF score for 18 language pairs of the IWSLT 2017. The blue-yellow scale is fit globally to the values across the table.}\label{tab:chrf}

\end{table}

As a second downstream task, we evaluate our segmentation on machine translation (MT) in a simulated low-resource setup.

\paragraph{Experimental setup.} We use the IWLST 2017 dataset of 18 language pairs (involving combinations of Arabic, English, Dutch, German, Italian, and Romanian)  with the provided data splits for train, validation, and testing. The exact language pairs and dataset statistics are in the Appendix in Table~\ref{tab:iwslt_data}. Similarly to POS tagging, we experiment with word-like and Morfessor pre-tokenization, BPE, and Unigram vocabulary construction (jointly on parallel data) and compare the default segmentation (Orig.) algorithms with the bigram-based segmentation distilled from the embedding-based segmentation algorithm (Ours).

We use the Transformer Base model \citep{vaswani-etal-2017-attention} as implemented in Marian \citep{junczys-dowmunt-etal-2018-marian}. We train the models using the Adam optimizer with learning rate $10^{-4}$ and the inverse square learning rate decay with 4,000 warmup steps with effective batch size 18,000 tokens.

\paragraph{Results.} We evaluate the MT quality using the chrF scores \citep{popovic-2015-chrf},\punctfootnote{We use the SacreBLEU implementation \citep{post-2018-call}: \texttt{chrF2|nrefs:1|case:mixed|eff:yes|nc:6|nw:0|\\space:no|version:2.0.0}} see Table~\ref{tab:chrf_all} in the Appendix for complete results.
At first glance, there are only minor differences in translation quality across the tested methods and language pairs, except for a few outliers.
Therefore, in Table~\ref{tab:chrf},\punctfootnote{Table~\ref{tab:chrf} shows normalized chrF scores. See Table~\ref{tab:bleu} in the Appendix for BLEU scores.} we provide aggregated results across the languages:
We first compute the mean chrF score per language pair and subtract it from the scores. Finally, we average the difference from the mean across languages. The results show that the word-based pre-tokenization outperforms Morfessor tokenization. Whilst our techniques have a slightly negative effect with the word-like pre-tokenization, adding Morfessor-based pre-tokenization shows significant improvements. Still, the overall MT quality stays behind the full Unigram and BPE preprocessing pipelines.


\begin{table}

\newcommand{\pearson}[2]{\gradientcell{#1}{-0.7}{0.7}{Cyan}{Yellow}{60}{#2}}

\setlength{\tabcolsep}{7pt}
\centering\footnotesize
\begin{tabular}{ccccccc@{\hskip14pt}c}
\toprule
cs & en & es & fr & hu & it & ru & \bf Avg. \\ \midrule
\pearson{-.30}{} &
\pearson{ .73}{\hphantom{-}} &
\pearson{ .69}{\hphantom{-}} &
\pearson{ .60}{\hphantom{-}} &
\pearson{ .95}{\hphantom{-}} &
\pearson{ .74}{\hphantom{-}} &
\pearson{ .33}{\hphantom{-}} &
\pearson{ .54}{\bf} \\
\bottomrule
\end{tabular}

\vspace{2pt}
(a) POS-tagging (accuracy)

\vspace{\baselineskip}

\setlength{\tabcolsep}{4pt}
\begin{tabular}{lc@{\hskip12pt} lc@{\hskip12pt} lc@{\hskip12pt} lc}
\toprule
ar $\ar$ en & \pearson{-.70}{} &
en $\ar$ ar & \pearson{-.73}{} &
de $\ar$ en & \pearson{-.13}{} &
en $\ar$ de & \pearson{-.38}{} \\

en $\ar$ fr & \pearson{-.06}{} & 
fr $\ar$ en & \pearson{ .19}{\hphantom{-}} &
en $\ar$ nl & \pearson{-.47}{} &
nl $\ar$ en & \pearson{ .03}{\hphantom{-}} \\

en $\ar$ ro & \pearson{-.31}{} &
ro $\ar$ en & \pearson{-.36}{} &
it $\ar$ en & \pearson{-.39}{} &
en $\ar$ it & \pearson{-.46}{} \\

it $\ar$ nl & \pearson{-.43}{} &
nl $\ar$ it & \pearson{-.40}{} &
ro $\ar$ it & \pearson{-.04}{} &
it $\ar$ ro & \pearson{-.17}{} \\

ro $\ar$ nl & \pearson{-.42}{} &
nl $\ar$ ro & \pearson{-.40}{} &
\multicolumn{2}{c}{$\Longrightarrow$} &
\bf Avg. &    \pearson{-.31}{\bf} \\

\bottomrule

\end{tabular}

\vspace{2pt}
(b) Machine Translation (chrF)

\caption{Pearson correlation of Rényi efficiency of the training data with the downstream performance. The blue-yellow scale is fit globally to the values across both tables.}\label{tab:renyi}

\end{table}

\subsection{Rényi Efficiency}

Finally, we evaluate the correlation between the results of our downstream tasks and Rényi Efficiency. \citet{zouhar-etal-2023-tokenization} conducted a theoretical analysis of information-theoretical properties of tokenizers and suggest to measure their unigram information efficiency. 
Information efficiency is the ratio of the unigram entropy of tokenized text and the maximum possible entropy given the vocabulary size. Instead of using the more common Shannon entropy, they use parametrized Rényi entropy with $\alpha=2.5$ that they claim better correlates with the downstream performance on English-German MT.

To verify the claims of \citet{zouhar-etal-2023-tokenization}, we computed the Pearson correlation of the Rényi efficiency of the training data in our experiments with the model performance. Our results are presented in Table~\ref{tab:renyi}. For POS tagging, Rényi efficiency is a good predictor of tagger performance in most languages except Czech. However, the correlation varies strongly between languages. In MT, we did not confirm the results of \citet{zouhar-etal-2023-tokenization}: the correlation of the Rényi efficiency of the training data and the MT quality in terms of chrF is mostly negative and highly varies across language pairs.

\section{Related Work}%
\label{sec:related}

\paragraph{Subword embeddings.}

There are relatively few methods for obtaining static subword embeddings. FastText \citep{bojanowski-etal-2017-enriching} averages subword embeddings to obtain static word embeddings. However, subwords are stored in a hash table with many conflicts for better memory efficiency, making the subword embeddings unusable for our purposes.
\citet{heinzerling-strube-2018-bpemb} trained subword embedding for 275 languages and various vocabulary sizes using GloVe \citep{pennington-etal-2014-glove} while treating subwords as standalone tokens. They, however, do not put the subword embeddings into relation to word embeddings. Static subword embeddings are, as the first layer, a part of most neural NLP models. However, none of the methods explicitly models the relationship between the words and subwords.

\paragraph{Subword segmentation.}

Besides the standard BPE \citep{sennrich-etal-2016-neural} and the Unigram model \citep{kudo-2018-subword}, several more recent approaches to subword segmentation exist.
\citet{xu-etal-2021-vocabulary} use optimal transport to find a replacement for greedy vocabulary construction of BPE, leading to more efficient bilingual vocabularies.
\citet{he-etal-2020-dynamic} and \citet{meyer-buys-2023-subword} work with Dynamic Programming Encoding that includes subword selection into the language-modeling objective of in MT model with a decoder using character-level inputs.
\citet{yehezkel-pinter-2023-incorporating} introduce SaGe, which uses skip-gram training objective as a loss to replace unigram perplexity used in the Unigram model. 
\citet{hofmann-etal-2022-embarrassingly} show that changing the segmentation algorithm in a WordPiece \citep{schuster2021japanese} tokenizer and a trained BERT model can improve classification performance.
\citet{schmidt2024tokenizaion} further elaborate on this idea and introduce an alternative segmentation algorithm that produces the minimum number of tokens given a vocabulary.

\section{Conclusions}

In this paper, we devised morphologically plausible methods for subword segmentation.
Inspired by \citet{schmidt2024tokenizaion}, we divide the tokenization process into three steps: pre-tokenization, vocabulary construction, and segmentation.

We described three key contributions of our work.
Our first contribution focuses on the pre-tokenization step:
Instead of the standard approaches, which split the text into word-like units, we use Morfessor, which splits the text into morphemes. However, we only regard this as pre-tokenization.
Next, we proposed a novel segmentation algorithm based on word and subword embeddings, which provides lexical grounding to the segmentation.
Finally, we proposed a statistical bigram segmentation model that can be used to simplify complex tokenization pipelines.

The intrinsic evaluation results show that the proposed method better captures language morphology than standard statistical subword segmentation approaches. This is further confirmed by the results we obtained on POS tagging, in which information about morphology is a key feature.

However, our method did not significantly improve the performance of machine translation, which is a more complex NLP task. 
We argue that a dedicated analysis would be required to determine the exact influence of the lexically grounded segmentation on the translation quality, which might be improved in one dimension but reduced in another.

In our work, we have taken steps to create a more morphologically accurate tokenization method while keeping the benefits of statistical subword segmentation. We believe these methods will improve modeling language overall and contribute to model interpretability and cross-lingual transfer.

\section{Limitations} 

The subword embedding formula derived in Section \ref{subsec:embeddings} requires a trained word embedding model and, therefore, relies on the quality of available data. This problem manifests mostly in under-represented languages, many of which would benefit from morphology-aware segmentation.

In Section \ref{sec:evaluation}, we use a set of heuristic rules to map the morphemes to the surface form for some languages. These rules are language agnostic and may introduce noise into the evaluation. However, the results are consistent with Czech, annotated on the morph level.


\section*{Acknowledgements}

We thank Tomasz Limisiewicz and Zdeněk Žabokrtský for discussing our early ideas on using subword embeddings for tokenization.

\noindent This work was supported by the Charles University project PRIMUS/23/SCI/023.

\bibliography{anthology,custom}

\appendix

\section{Code Examples}
\label{app:code}

Below, we list Python implementations of the two proposed segmentation algorithms: Segmentation based on subword embeddings (Algorithm~\ref{alg:embedding}) and bigram segmentation (Algorithm~\ref{alg:bigram}).

\begin{lstfloat*}
\begin{lstlisting}[language=Python]  
def embedding_segment(
        word: str,
        word_embedding: np.ndarray,
        subword_embeddings: Dict[str, np.ndarray]) -> List[str]:

    # Costs of segmenting the word up to a certain length
    costs = [0. for _ in range(len(word) + 1)]
    # Backward pointers: position i says from what index we can get position i
    prev = [0 for _ in word]

    # 1. Populate the segmentation cost table
    for i in range(1, len(word) + 1):
        # Now, we know how to segment everything up to position i-1 and want to find position i
        indices = []  # Indices j from where we can go to position i
        scores = []   # Scores corresponding to the indices
        for j in range(i):  # 0..i
            subword = word[j:i]
            if subword in subword_embeddings:
                subword_embedding = subword_embeddings[subword]
                new_cost = costs[j] + cosine_similarity(
                    word_embedding, subword_embedding) - 1
                scores.append(new_cost)
                indices.append(j)
        # Best index from which we get to position i, i.e., the argmax of scores
        idx = max(range(len(scores)), key=lambda i: scores[i])
        costs[i] = scores[idx]
        prev[i - 1] = indices[idx]

    # 2. Reconstruct the best segmentation by following the backward pointers
    subwords = []
    idx = len(prev) - 1
    while idx >= 0:
        new_idx = prev[idx]
        sbwrd = word[new_idx:idx + 1]
        subwords.append(sbwrd)

        idx = new_idx - 1
    return list(reversed(subwords)), costs[-1]
\end{lstlisting}
\caption{Python code showing the segmentation algorithm using subword embeddings. On the input, \texttt{word} is the word to be segmented, \texttt{word\_embedding} is its embedding, and \texttt{subword\_embedding} is the subword embedding matrix. \\[1ex] It is a dynamic programming algorithm that first computes the scores of the best segmentation up to a given position in the string (kept in list \texttt{costs}) and what was the start index of the last subword in the best-scoring segmentation (kept in list \texttt{prev}). When moving to the next index in the for loop on line 12, we can rely on knowing the best segmentation score for all indices up to $i-1$ from the previous iteration. Therefore, in the for loop on line 16, we can try all subwords that will bring us to index $i$. figure out the best possible subword that will extend the segmentation to index $i$. \\[1ex] In the second step, we use the list \texttt{prev} to reconstruct what subwords were used to the best score starting at the end of the word.}%
\label{alg:embedding}
\end{lstfloat*}

\begin{lstfloat*}
\begin{lstlisting}[language=Python]
def beam_search_segment(
        word: str,
        vocab: Set[str]
        beam_size: int = 5) -> List[str]:
    max_subword_length = max(len(tok) for tok in vocab)

    # List where the i-th position contains possible segmentations ending at position i
    segmentations = [[(["###"], 0.0)]] + [[] for _ in token]
    for start in range(len(token)):
        # Try to expand all segmentations ending at index `start`
        # with subwords of all possible lengths
        for length in range(1, vocab.max_subword_length + 1):
            end = start + length
            if end > len(token):
                break

            subword = token[start:end]
            if subword not in vocab and len(subword) > 1:
                continue

            # Expand from the current segmentations ending at index `start`
            for prev_segmentation, prev_score in segmentations[start]:
                # Compute the bigram log probability of the current `subword`
                # given the last subword of `prev_segmenation`
                score = log_probability(subword, prev_segmentation[-1])
                new_segmentation = prev_segmentation + [subword]
                new_score = prev_score + score  # Summing log probabilities
                segmentations[end].append((new_segmentation, new_score))

        # For each end index that follows, keep only the best `beam_size` segmentations
        for i, seg_list in enumerate(segmentations[start + 1:]):
            if len(seg_list) > beam_size:
                seg_list.sort(key=lambda x: x[1], reverse=True)
                segmentations[start + 1 + i] = seg_list[:beam_size]

    best_segmentation = max(segmentations[-1], key=lambda x: x[1])
    return best_segmentation[0][1:]
\end{lstlisting}
\caption{Python code for bigram segmentation. On the input, \texttt{token} is the token to be tokenized, \texttt{vocab} is the subword vocabulary, and \texttt{max\_subword\_length} controls the maximum number of characters in a subword. We assume there is a function \texttt{log\_probability} that computes the log probability of a subword bigram. }\label{alg:bigram} \end{lstfloat*}

\section{Statistis of Used Datasets}

Table~\ref{tab:ud_data} contains statistics of the UD Treebanks used for POS Tagging evaluation. Table~\ref{tab:iwslt_data} contains basic statics of the IWSLT 2017 data used for machine translation evaluation.

\begin{table}[ht]

\centering \footnotesize
\newcommand{\Chrfmean}[2]{\gradientcell{#1}{-1.526572344768285}{1.4373875527510656}{Cyan}{Yellow}{60}{#2}}
\begin{tabular}{lll ccc@{\hskip 18pt}c}
\toprule
\multicolumn{3}{l}{\multirow{2}{*}{Tokenization}} & \multicolumn{3}{c}{Vocabulary\hspace*{15pt}} & \multirow{2}{*}{Avg.} \\ \cmidrule(lr{18pt}){4-6}
& & & 4k & 8k & 16k & \\ \midrule

\multirow{4}{*}{\rotatebox[origin=c]{90}{Word-like}} & \multirow{2}{*}{BPE} & Orig. & \Chrfmean{0.3}{\hphantom{-}} & \Chrfmean{0.7}{\hphantom{-}} & \Chrfmean{0.7}{\hphantom{-}} & \Chrfmean{0.5}{\hphantom{-}} \\
& & Ours & \Chrfmean{0.3}{\hphantom{-}} & \Chrfmean{0.4}{\hphantom{-}} & \Chrfmean{0.5}{\hphantom{-}} & \Chrfmean{0.4}{\hphantom{-}} \\ \cmidrule{2-7}
& \multirow{2}{*}{Uni.} & Orig. & \Chrfmean{0.3}{\hphantom{-}} & \Chrfmean{0.9}{\hphantom{-}} & \Chrfmean{0.7}{\hphantom{-}} & \Chrfmean{0.7}{\hphantom{-}} \\
& & Ours & \Chrfmean{0.2}{\hphantom{-}} & \Chrfmean{0.8}{\hphantom{-}} & \Chrfmean{0.3}{\hphantom{-}} & \Chrfmean{0.5}{\hphantom{-}} \\ \midrule
\multirow{4}{*}{\rotatebox[origin=c]{90}{Morfessor}} & \multirow{2}{*}{BPE} & Orig. & \Chrfmean{-0.7}{} & \Chrfmean{-0.7}{} & \Chrfmean{-0.8}{} & \Chrfmean{-0.7}{} \\
& & Ours & \Chrfmean{-0.4}{} & \Chrfmean{-0.1}{} & \Chrfmean{-0.0}{} & \Chrfmean{-0.2}{} \\ \cmidrule{2-7}
& \multirow{2}{*}{Uni.} & Orig. & \Chrfmean{-1.0}{} & \Chrfmean{-0.8}{} & \Chrfmean{-1.0}{} & \Chrfmean{-0.9}{} \\
& & Ours & \Chrfmean{-0.1}{} & \Chrfmean{-0.0}{} & \Chrfmean{-0.5}{} & \Chrfmean{-0.2}{} \\ \bottomrule
\end{tabular}

\caption{Mean deviation from the average BLEU score for 18 language pairs of the IWSLT 2017. The blue-yellow scale is fit globally to the values across the table.}\label{tab:bleu}

\end{table}

\begin{table*}[ht]

\centering\footnotesize
\begin{tabular}{l cc cc cc}
\toprule
Treebank & \multicolumn{2}{c}{Train} & \multicolumn{2}{c}{Dev} & \multicolumn{2}{c}{Test} \\ \cmidrule(lr){2-3} \cmidrule(lr){4-5} \cmidrule(lr){6-7}
& Sent. & Tokens & Sent & Tokens & Sent. & Tokens \\ \midrule

Czech PDT &   68k & 1,192k &    9k  & 162k &   10k  & 177k \\
English EWT & 12k & 207k & 2k & 25k & 2k & 25k \\
Spanish GSD & 14k & 389k & 1k & 37k & 1k & 12k \\
French GSD & 14k & 364k & 1k & 36k & 1k & 10k \\
Hungarian Szeged & 1k & 20k & 1k & 11k & 1k & 10k \\
Italian ISDT & 13k & 294k & 1k & 12k & 1k & 11k \\
Russian SynTagRus & 69k & 1206k & 8k & 153k & 8k & 157 \\
\bottomrule

\end{tabular}

\caption{Basic statistics of the splits of the UD treebanks used in the POS tagging evaluation in terms of number sentences and number of tokens.}\label{tab:ud_data}

\end{table*}

\begin{table*}[ht]
\centering\footnotesize
\begin{tabular}{l ccc ccc ccc}
\toprule
Language pair & \multicolumn{3}{c}{Train} & \multicolumn{3}{c}{Dev} & \multicolumn{3}{c}{Test} \\ \cmidrule(lr){2-4} \cmidrule(lr){5-7} \cmidrule(lr){8-10}
& Sent & Src. tok & Tgt. tok. & Sent & Src. tok & Tgt. tok. & Sent & Src. tok & Tgt. tok. \\ \midrule
ar-en & 231k & 3,817k & 4,865k & 1k & 15k & 21k & 8k & 136k & 184k \\
de-en & 206k & 3,923k & 4,318k & 1k & 19k & 21k & 8k & 149k & 162k \\
en-fr & 232k & 4,888k & 5,360k & 1k & 21k & 21k & 8k & 184k & 193k \\
en-nl & 237k & 4,540k & 4,009k & 1k & 20k & 19k & 2k &  33k &  31k \\
en-ro & 220k & 4,594k & 4,201k & 1k & 20k & 20k & 2k &  33k &  32k \\
it-en & 231k & 4,846k & 4,450k & 1k & 20k & 19k & 2k &  32k &  31k \\
it-nl & 233k & 4,105k & 3,944k & 1k & 18k & 19k & 2k &  29k &  31k \\
ro-it & 217k & 4,169k & 4,148k & 1k & 18k & 20k & 2k &  29k &  31k \\
ro-nl & 206k & 3,809k & 3,939k & 1k & 18k & 20k & 2k &  30k &  32k \\
\bottomrule
\end{tabular}

\caption{Sizes of the IWSLT 2017 datasets in terms of the number of sentence pairs and the number of tokens on the source and the target side.}\label{tab:iwslt_data}

\end{table*}

\begin{table*}[ht]

\footnotesize\centering

\newcommand{\PrecisioncsTwo}[2]{\gradientcell{#1}{74.3}{90.1}{Cyan}{Yellow}{60}{#2}}
\newcommand{\RenyicsTwo}[2]{\gradientcell{#1}{0.43}{0.462}{Cyan}{Yellow}{60}{#2}}
\newcommand{\PrecisionenTwo}[2]{\gradientcell{#1}{54.7}{70.4}{Cyan}{Yellow}{60}{#2}}
\newcommand{\RenyienTwo}[2]{\gradientcell{#1}{0.435}{0.443}{Cyan}{Yellow}{60}{#2}}
\newcommand{\PrecisionesTwo}[2]{\gradientcell{#1}{59.4}{64.5}{Cyan}{Yellow}{60}{#2}}
\newcommand{\RenyiesTwo}[2]{\gradientcell{#1}{0.404}{0.431}{Cyan}{Yellow}{60}{#2}}
\newcommand{\PrecisionfrTwo}[2]{\gradientcell{#1}{55.3}{65.3}{Cyan}{Yellow}{60}{#2}}
\newcommand{\RenyifrTwo}[2]{\gradientcell{#1}{0.428}{0.456}{Cyan}{Yellow}{60}{#2}}
\newcommand{\PrecisionhuTwo}[2]{\gradientcell{#1}{75.3}{84.1}{Cyan}{Yellow}{60}{#2}}
\newcommand{\RenyihuTwo}[2]{\gradientcell{#1}{0.382}{0.409}{Cyan}{Yellow}{60}{#2}}
\newcommand{\PrecisionitTwo}[2]{\gradientcell{#1}{51.0}{60.9}{Cyan}{Yellow}{60}{#2}}
\newcommand{\RenyiitTwo}[2]{\gradientcell{#1}{0.444}{0.462}{Cyan}{Yellow}{60}{#2}}
\newcommand{\PrecisionmnTwo}[2]{\gradientcell{#1}{76.0}{90.9}{Cyan}{Yellow}{60}{#2}}
\newcommand{\RenyimnTwo}[2]{\gradientcell{#1}{0.474}{0.504}{Cyan}{Yellow}{60}{#2}}
\newcommand{\PrecisionruTwo}[2]{\gradientcell{#1}{60.7}{72.3}{Cyan}{Yellow}{60}{#2}}
\newcommand{\RenyiruTwo}[2]{\gradientcell{#1}{0.425}{0.464}{Cyan}{Yellow}{60}{#2}}

\begin{tabular}{lll cccccccc@{\hskip 18pt}cccccccc}
\toprule

\multicolumn{2}{l}{Vocab.} & Inf. & \multicolumn{8}{c}{Morpheme boundary precision} & \multicolumn{8}{c}{Rényi efficiency} \\ \cmidrule(lr{18pt}){4-11} \cmidrule(lr){12-19}
& & & cs & en & es & fr & hu & it & mn & ru &  cs & en & es & fr & hu & it & mn & ru \\ \midrule
\multirow{6}{*}{\rotatebox{90}{Word-like}} & \multirow{3}{*}{BPE} & Orig & \PrecisioncsTwo{74.3}{} & \PrecisionenTwo{54.7}{} & \PrecisionesTwo{59.4}{} & \PrecisionfrTwo{55.3}{} & \PrecisionhuTwo{75.3}{} & \PrecisionitTwo{51.0}{} & \PrecisionmnTwo{76.0}{} & \PrecisionruTwo{60.7}{} & \RenyicsTwo{.430}{} & \RenyienTwo{.435}{} & \RenyiesTwo{.404}{} & \RenyifrTwo{.428}{} & \RenyihuTwo{.382}{} & \RenyiitTwo{.444}{} & \RenyimnTwo{.478}{} & \RenyiruTwo{.425}{}\\
& & Emb. & \PrecisioncsTwo{77.2}{} & \PrecisionenTwo{63.2}{} & \PrecisionesTwo{62.8}{} & \PrecisionfrTwo{60.7}{} & \PrecisionhuTwo{80.5}{} & \PrecisionitTwo{56.4}{} & \PrecisionmnTwo{86.2}{} & \PrecisionruTwo{63.1}{} & \RenyicsTwo{.432}{} & \RenyienTwo{.441}{} & \RenyiesTwo{.410}{} & \RenyifrTwo{.434}{} & \RenyihuTwo{.396}{} & \RenyiitTwo{.450}{} & \RenyimnTwo{.487}{} & \RenyiruTwo{.434}{}\\
& & Big. & \PrecisioncsTwo{77.7}{} & \PrecisionenTwo{63.7}{} & \PrecisionesTwo{63.0}{} & \PrecisionfrTwo{60.5}{} & \PrecisionhuTwo{79.7}{} & \PrecisionitTwo{56.3}{} & \PrecisionmnTwo{85.8}{} & \PrecisionruTwo{64.9}{} & \RenyicsTwo{.433}{} & \RenyienTwo{.441}{} & \RenyiesTwo{.411}{} & \RenyifrTwo{.435}{} & \RenyihuTwo{.396}{} & \RenyiitTwo{.451}{} & \RenyimnTwo{.488}{} & \RenyiruTwo{.435}{}\\ \cmidrule(l){2-19}
& \multirow{3}{*}{Uni.} & Orig & \PrecisioncsTwo{82.9}{} & \PrecisionenTwo{62.2}{} & \PrecisionesTwo{61.2}{} & \PrecisionfrTwo{62.5}{} & \PrecisionhuTwo{78.5}{} & \PrecisionitTwo{51.2}{} & \PrecisionmnTwo{88.9}{} & \PrecisionruTwo{65.0}{} & \RenyicsTwo{.434}{} & \RenyienTwo{.437}{} & \RenyiesTwo{.404}{} & \RenyifrTwo{.430}{} & \RenyihuTwo{.391}{} & \RenyiitTwo{.448}{} & \RenyimnTwo{.484}{} & \RenyiruTwo{.432}{}\\
& & Emb. & \PrecisioncsTwo{85.7}{} & \PrecisionenTwo{66.4}{} & \PrecisionesTwo{63.4}{} & \PrecisionfrTwo{64.3}{} & \PrecisionhuTwo{80.8}{} & \PrecisionitTwo{55.6}{} & \PrecisionmnTwo{88.2}{} & \PrecisionruTwo{66.5}{} & \RenyicsTwo{.433}{} & \RenyienTwo{.443}{\bf } & \RenyiesTwo{.414}{} & \RenyifrTwo{.439}{} & \RenyihuTwo{.398}{} & \RenyiitTwo{.453}{} & \RenyimnTwo{.474}{} & \RenyiruTwo{.439}{}\\
& & Big. & \PrecisioncsTwo{85.5}{} & \PrecisionenTwo{66.7}{} & \PrecisionesTwo{63.3}{} & \PrecisionfrTwo{64.4}{} & \PrecisionhuTwo{80.6}{} & \PrecisionitTwo{55.6}{} & \PrecisionmnTwo{87.9}{} & \PrecisionruTwo{67.9}{} & \RenyicsTwo{.435}{} & \RenyienTwo{.443}{\bf } & \RenyiesTwo{.415}{} & \RenyifrTwo{.440}{} & \RenyihuTwo{.399}{} & \RenyiitTwo{.454}{} & \RenyimnTwo{.475}{} & \RenyiruTwo{.442}{}\\ \midrule
\multirow{6}{*}{\rotatebox{90}{Morfessor}} & \multirow{3}{*}{BPE} & Orig & \PrecisioncsTwo{86.8}{} & \PrecisionenTwo{68.3}{} & \PrecisionesTwo{64.5}{\bf } & \PrecisionfrTwo{64.2}{} & \PrecisionhuTwo{80.0}{} & \PrecisionitTwo{60.1}{} & \PrecisionmnTwo{88.6}{} & \PrecisionruTwo{67.4}{} & \RenyicsTwo{.454}{} & \RenyienTwo{.440}{} & \RenyiesTwo{.425}{} & \RenyifrTwo{.449}{} & \RenyihuTwo{.398}{} & \RenyiitTwo{.458}{} & \RenyimnTwo{.499}{} & \RenyiruTwo{.455}{}\\
& & Emb. & \PrecisioncsTwo{87.9}{} & \PrecisionenTwo{70.4}{\bf } & \PrecisionesTwo{64.4}{} & \PrecisionfrTwo{65.1}{} & \PrecisionhuTwo{83.1}{} & \PrecisionitTwo{59.7}{} & \PrecisionmnTwo{90.7}{} & \PrecisionruTwo{69.9}{} & \RenyicsTwo{.456}{} & \RenyienTwo{.443}{\bf } & \RenyiesTwo{.428}{} & \RenyifrTwo{.452}{} & \RenyihuTwo{.407}{} & \RenyiitTwo{.460}{} & \RenyimnTwo{.502}{} & \RenyiruTwo{.460}{}\\
& & Big. & \PrecisioncsTwo{87.6}{} & \PrecisionenTwo{67.9}{} & \PrecisionesTwo{64.2}{} & \PrecisionfrTwo{65.3}{\bf } & \PrecisionhuTwo{82.5}{} & \PrecisionitTwo{60.3}{} & \PrecisionmnTwo{90.2}{} & \PrecisionruTwo{69.6}{} & \RenyicsTwo{.457}{} & \RenyienTwo{.443}{\bf } & \RenyiesTwo{.429}{} & \RenyifrTwo{.452}{} & \RenyihuTwo{.407}{} & \RenyiitTwo{.461}{} & \RenyimnTwo{.503}{} & \RenyiruTwo{.461}{}\\ \cmidrule(l){2-19}
& \multirow{3}{*}{Uni.} & Orig & \PrecisioncsTwo{88.3}{} & \PrecisionenTwo{68.8}{} & \PrecisionesTwo{63.9}{} & \PrecisionfrTwo{64.1}{} & \PrecisionhuTwo{82.1}{} & \PrecisionitTwo{60.1}{} & \PrecisionmnTwo{89.9}{} & \PrecisionruTwo{68.9}{} & \RenyicsTwo{.461}{} & \RenyienTwo{.442}{} & \RenyiesTwo{.427}{} & \RenyifrTwo{.453}{} & \RenyihuTwo{.404}{} & \RenyiitTwo{.462}{\bf } & \RenyimnTwo{.504}{\bf } & \RenyiruTwo{.464}{\bf }\\
& & Emb. & \PrecisioncsTwo{90.1}{\bf } & \PrecisionenTwo{69.0}{} & \PrecisionesTwo{64.4}{} & \PrecisionfrTwo{65.0}{} & \PrecisionhuTwo{84.1}{\bf } & \PrecisionitTwo{60.9}{\bf } & \PrecisionmnTwo{90.9}{\bf } & \PrecisionruTwo{72.3}{\bf } & \RenyicsTwo{.461}{} & \RenyienTwo{.443}{\bf } & \RenyiesTwo{.430}{} & \RenyifrTwo{.456}{\bf } & \RenyihuTwo{.409}{\bf } & \RenyiitTwo{.461}{} & \RenyimnTwo{.497}{} & \RenyiruTwo{.461}{}\\
& & Big. & \PrecisioncsTwo{89.2}{} & \PrecisionenTwo{68.1}{} & \PrecisionesTwo{63.9}{} & \PrecisionfrTwo{65.2}{} & \PrecisionhuTwo{82.8}{} & \PrecisionitTwo{60.4}{} & \PrecisionmnTwo{90.4}{} & \PrecisionruTwo{70.7}{} & \RenyicsTwo{.462}{\bf } & \RenyienTwo{.443}{\bf } & \RenyiesTwo{.431}{\bf } & \RenyifrTwo{.456}{\bf } & \RenyihuTwo{.409}{\bf } & \RenyiitTwo{.462}{\bf } & \RenyimnTwo{.497}{} & \RenyiruTwo{.463}{}\\

\bottomrule
\end{tabular}

\newcommand{\PrecisioncsFour}[2]{\gradientcell{#1}{78.1}{91.4}{Cyan}{Yellow}{60}{#2}}
\newcommand{\RenyicsFour}[2]{\gradientcell{#1}{0.412}{0.457}{Cyan}{Yellow}{60}{#2}}
\newcommand{\PrecisionenFour}[2]{\gradientcell{#1}{58.1}{73.5}{Cyan}{Yellow}{60}{#2}}
\newcommand{\RenyienFour}[2]{\gradientcell{#1}{0.425}{0.441}{Cyan}{Yellow}{60}{#2}}
\newcommand{\PrecisionesFour}[2]{\gradientcell{#1}{61.4}{69.1}{Cyan}{Yellow}{60}{#2}}
\newcommand{\RenyiesFour}[2]{\gradientcell{#1}{0.392}{0.429}{Cyan}{Yellow}{60}{#2}}
\newcommand{\PrecisionfrFour}[2]{\gradientcell{#1}{58.6}{69.7}{Cyan}{Yellow}{60}{#2}}
\newcommand{\RenyifrFour}[2]{\gradientcell{#1}{0.417}{0.454}{Cyan}{Yellow}{60}{#2}}
\newcommand{\PrecisionhuFour}[2]{\gradientcell{#1}{78.1}{86.9}{Cyan}{Yellow}{60}{#2}}
\newcommand{\RenyihuFour}[2]{\gradientcell{#1}{0.366}{0.4}{Cyan}{Yellow}{60}{#2}}
\newcommand{\PrecisionitFour}[2]{\gradientcell{#1}{54.6}{66.8}{Cyan}{Yellow}{60}{#2}}
\newcommand{\RenyiitFour}[2]{\gradientcell{#1}{0.433}{0.459}{Cyan}{Yellow}{60}{#2}}
\newcommand{\PrecisionmnFour}[2]{\gradientcell{#1}{80.3}{93.0}{Cyan}{Yellow}{60}{#2}}
\newcommand{\RenyimnFour}[2]{\gradientcell{#1}{0.464}{0.499}{Cyan}{Yellow}{60}{#2}}
\newcommand{\PrecisionruFour}[2]{\gradientcell{#1}{62.6}{73.7}{Cyan}{Yellow}{60}{#2}}
\newcommand{\RenyiruFour}[2]{\gradientcell{#1}{0.407}{0.46}{Cyan}{Yellow}{60}{#2}}

\begin{tabular}{lll cccccccc@{\hskip 18pt}cccccccc}
\toprule

\multicolumn{2}{l}{Vocab.} & Inf. & \multicolumn{8}{c}{Morpheme boundary precision} & \multicolumn{8}{c}{Rényi efficiency} \\ \cmidrule(lr{18pt}){4-11} \cmidrule(lr){12-19}
& & & cs & en & es & fr & hu & it & mn & ru &  cs & en & es & fr & hu & it & mn & ru \\ \midrule
\multirow{6}{*}{\rotatebox{90}{Word-like}} & \multirow{3}{*}{BPE} & Orig & \PrecisioncsFour{78.1}{} & \PrecisionenFour{58.1}{} & \PrecisionesFour{61.4}{} & \PrecisionfrFour{58.6}{} & \PrecisionhuFour{78.1}{} & \PrecisionitFour{54.6}{} & \PrecisionmnFour{80.3}{} & \PrecisionruFour{62.6}{} & \RenyicsFour{.412}{} & \RenyienFour{.425}{} & \RenyiesFour{.392}{} & \RenyifrFour{.417}{} & \RenyihuFour{.366}{} & \RenyiitFour{.433}{} & \RenyimnFour{.466}{} & \RenyiruFour{.407}{}\\
& & Emb. & \PrecisioncsFour{81.3}{} & \PrecisionenFour{67.2}{} & \PrecisionesFour{65.1}{} & \PrecisionfrFour{65.2}{} & \PrecisionhuFour{83.5}{} & \PrecisionitFour{60.1}{} & \PrecisionmnFour{90.4}{} & \PrecisionruFour{64.8}{} & \RenyicsFour{.415}{} & \RenyienFour{.431}{} & \RenyiesFour{.400}{} & \RenyifrFour{.424}{} & \RenyihuFour{.381}{} & \RenyiitFour{.439}{} & \RenyimnFour{.475}{} & \RenyiruFour{.418}{}\\
& & Big. & \PrecisioncsFour{81.8}{} & \PrecisionenFour{68.7}{} & \PrecisionesFour{65.4}{} & \PrecisionfrFour{66.3}{} & \PrecisionhuFour{83.4}{} & \PrecisionitFour{61.0}{} & \PrecisionmnFour{90.4}{} & \PrecisionruFour{67.7}{} & \RenyicsFour{.416}{} & \RenyienFour{.431}{} & \RenyiesFour{.401}{} & \RenyifrFour{.424}{} & \RenyihuFour{.382}{} & \RenyiitFour{.440}{} & \RenyimnFour{.476}{} & \RenyiruFour{.419}{}\\ \cmidrule(l){2-19}
& \multirow{3}{*}{Uni.} & Orig & \PrecisioncsFour{85.6}{} & \PrecisionenFour{66.7}{} & \PrecisionesFour{65.0}{} & \PrecisionfrFour{66.9}{} & \PrecisionhuFour{81.8}{} & \PrecisionitFour{55.1}{} & \PrecisionmnFour{92.0}{} & \PrecisionruFour{67.8}{} & \RenyicsFour{.417}{} & \RenyienFour{.429}{} & \RenyiesFour{.394}{} & \RenyifrFour{.421}{} & \RenyihuFour{.376}{} & \RenyiitFour{.438}{} & \RenyimnFour{.474}{} & \RenyiruFour{.417}{}\\
& & Emb. & \PrecisioncsFour{87.9}{} & \PrecisionenFour{69.9}{} & \PrecisionesFour{66.7}{} & \PrecisionfrFour{68.1}{} & \PrecisionhuFour{83.7}{} & \PrecisionitFour{58.7}{} & \PrecisionmnFour{91.5}{} & \PrecisionruFour{68.4}{} & \RenyicsFour{.418}{} & \RenyienFour{.434}{} & \RenyiesFour{.404}{} & \RenyifrFour{.430}{} & \RenyihuFour{.384}{} & \RenyiitFour{.443}{} & \RenyimnFour{.464}{} & \RenyiruFour{.426}{}\\
& & Big. & \PrecisioncsFour{88.2}{} & \PrecisionenFour{71.0}{} & \PrecisionesFour{66.4}{} & \PrecisionfrFour{69.1}{} & \PrecisionhuFour{83.9}{} & \PrecisionitFour{60.5}{} & \PrecisionmnFour{91.6}{} & \PrecisionruFour{70.9}{} & \RenyicsFour{.420}{} & \RenyienFour{.434}{} & \RenyiesFour{.405}{} & \RenyifrFour{.430}{} & \RenyihuFour{.385}{} & \RenyiitFour{.444}{} & \RenyimnFour{.465}{} & \RenyiruFour{.427}{}\\ \midrule
\multirow{6}{*}{\rotatebox{90}{Morfessor}} & \multirow{3}{*}{BPE} & Orig & \PrecisioncsFour{90.6}{} & \PrecisionenFour{73.5}{\bf } & \PrecisionesFour{69.1}{\bf } & \PrecisionfrFour{69.7}{\bf } & \PrecisionhuFour{85.4}{} & \PrecisionitFour{66.8}{\bf } & \PrecisionmnFour{92.5}{} & \PrecisionruFour{71.8}{} & \RenyicsFour{.446}{} & \RenyienFour{.436}{} & \RenyiesFour{.421}{} & \RenyifrFour{.446}{} & \RenyihuFour{.388}{} & \RenyiitFour{.454}{} & \RenyimnFour{.497}{} & \RenyiruFour{.448}{}\\
& & Emb. & \PrecisioncsFour{89.9}{} & \PrecisionenFour{72.6}{} & \PrecisionesFour{67.3}{} & \PrecisionfrFour{68.5}{} & \PrecisionhuFour{86.3}{} & \PrecisionitFour{63.6}{} & \PrecisionmnFour{93.0}{\bf } & \PrecisionruFour{72.6}{} & \RenyicsFour{.449}{} & \RenyienFour{.439}{} & \RenyiesFour{.424}{} & \RenyifrFour{.448}{} & \RenyihuFour{.397}{} & \RenyiitFour{.456}{} & \RenyimnFour{.499}{\bf } & \RenyiruFour{.453}{}\\
& & Big. & \PrecisioncsFour{89.4}{} & \PrecisionenFour{71.4}{} & \PrecisionesFour{67.1}{} & \PrecisionfrFour{68.6}{} & \PrecisionhuFour{85.5}{} & \PrecisionitFour{64.6}{} & \PrecisionmnFour{92.5}{} & \PrecisionruFour{72.5}{} & \RenyicsFour{.449}{} & \RenyienFour{.439}{} & \RenyiesFour{.424}{} & \RenyifrFour{.448}{} & \RenyihuFour{.396}{} & \RenyiitFour{.456}{} & \RenyimnFour{.499}{\bf } & \RenyiruFour{.454}{}\\ \cmidrule(l){2-19}
& \multirow{3}{*}{Uni.} & Orig & \PrecisioncsFour{90.2}{} & \PrecisionenFour{70.8}{} & \PrecisionesFour{65.7}{} & \PrecisionfrFour{65.8}{} & \PrecisionhuFour{85.4}{} & \PrecisionitFour{63.3}{} & \PrecisionmnFour{91.9}{} & \PrecisionruFour{71.6}{} & \RenyicsFour{.456}{} & \RenyienFour{.441}{\bf } & \RenyiesFour{.426}{} & \RenyifrFour{.452}{} & \RenyihuFour{.395}{} & \RenyiitFour{.452}{} & \RenyimnFour{.491}{} & \RenyiruFour{.460}{\bf }\\
& & Emb. & \PrecisioncsFour{91.4}{\bf } & \PrecisionenFour{70.4}{} & \PrecisionesFour{65.0}{} & \PrecisionfrFour{65.7}{} & \PrecisionhuFour{86.9}{\bf } & \PrecisionitFour{63.3}{} & \PrecisionmnFour{92.7}{} & \PrecisionruFour{73.7}{\bf } & \RenyicsFour{.456}{} & \RenyienFour{.441}{\bf } & \RenyiesFour{.428}{} & \RenyifrFour{.454}{\bf } & \RenyihuFour{.400}{\bf } & \RenyiitFour{.458}{} & \RenyimnFour{.492}{} & \RenyiruFour{.456}{}\\
& & Big. & \PrecisioncsFour{90.8}{} & \PrecisionenFour{70.0}{} & \PrecisionesFour{65.5}{} & \PrecisionfrFour{66.4}{} & \PrecisionhuFour{86.1}{} & \PrecisionitFour{63.2}{} & \PrecisionmnFour{92.1}{} & \PrecisionruFour{73.1}{} & \RenyicsFour{.457}{\bf } & \RenyienFour{.441}{\bf } & \RenyiesFour{.429}{\bf } & \RenyifrFour{.454}{\bf } & \RenyihuFour{.400}{\bf } & \RenyiitFour{.459}{\bf } & \RenyimnFour{.492}{} & \RenyiruFour{.459}{}\\

\bottomrule
\end{tabular}

\caption{Morpheme boundary precision on the SIGMORPHON 2018 test set and Rényi efficiency estimated on 4k plain text sentences for tokenizers with 24k and 40k-sized vocabularies. The best results in each column are bolded. The blue-yellow scale is fit to the value range per column.}\label{tab:intrinsicAppendix}

\end{table*}

\begin{table*}[ht]
\resizebox{\textwidth}{!}{%
\centering\footnotesize
\newcommand{\Aggr}[2]{\gradientcell{#1}{-2.013159192681013}{0.7453248272368196}{Cyan}{Yellow}{60}{#2}}
\newcolumntype{C}{>{\centering}m{15pt}}\begin{tabular}{lll ccccccc@{\hskip 18pt}c}
\toprule
\multicolumn{3}{l}{Tokenization} &cs & en & es & fr & hu & it & ru & Aggr.\\ \midrule
\multicolumn{3}{l}{Most frequent unigram} & 91.70 & 83.30 & 88.00 & 89.60 & 60.40 & 90.30 & 88.80 & \\
\multicolumn{3}{l}{HMM Tagger} & 93.70 & 87.60 & 91.70 & 93.00 & 72.80 & 93.30 & 91.00 & \\ \midrule
\multicolumn{3}{l}{Word vocab} &  96.16 \scriptsize (0.19) & 92.07 \scriptsize (0.56) & 94.43 \scriptsize (0.24) & 96.14 \scriptsize (0.30) & 79.44 \scriptsize (1.10) & 96.45 \scriptsize (0.27) & 94.16 \scriptsize (0.51) & \Aggr{-2.013}{} \\
\multicolumn{3}{l}{Morfessor} &  96.01 \scriptsize (0.35) & 92.05 \scriptsize (0.64) & 94.61 \scriptsize (0.22) & 96.19 \scriptsize (0.24) & 78.14 \scriptsize (1.86) & 96.64 \scriptsize (0.15) & 94.48 \scriptsize (0.45) & \Aggr{-1.902}{} \\ \midrule
\multirow{4}{*}{\rotatebox[origin=c]{90}{Word-like}} & \multirow{2}{*}{BPE} & Orig. &  98.17 \scriptsize (0.03) & 93.73 \scriptsize (0.16) & 95.50 \scriptsize (0.14) & 97.16 \scriptsize (0.08) & 87.76 \scriptsize (1.02) & 97.47 \scriptsize (0.10) & 97.38 \scriptsize (0.05) & \Aggr{0.340}{\hphantom{-}} \\
& & Ours &  98.19 \scriptsize (0.03) & 93.78 \scriptsize (0.15) & 95.58 \scriptsize (0.09) & 97.23 \scriptsize (0.12) & 88.88 \scriptsize (0.92) & 97.56 \scriptsize (0.07) & 97.40 \scriptsize (0.03) & \Aggr{0.471}{\hphantom{-}} \\ \cmidrule{2-11}
& \multirow{2}{*}{Uni.} & Orig. &  98.09 \scriptsize (0.08) & 93.50 \scriptsize (0.17) & 95.41 \scriptsize (0.09) & 97.00 \scriptsize (0.06) & 88.57 \scriptsize (0.50) & 97.41 \scriptsize (0.10) & 97.30 \scriptsize (0.04) & \Aggr{0.187}{\hphantom{-}} \\
& & Ours &  98.17 \scriptsize (0.04) & 93.76 \scriptsize (0.20) & 95.56 \scriptsize (0.11) & 97.11 \scriptsize (0.12) & 89.68 \scriptsize (0.50) & 97.58 \scriptsize (0.09) & 97.43 \scriptsize (0.05) & \Aggr{0.447}{\hphantom{-}} \\ \midrule
\multirow{4}{*}{\rotatebox[origin=c]{90}{Morfessor}} & \multirow{2}{*}{BPE} & Orig. &  98.18 \scriptsize (0.02) & 93.91 \scriptsize (0.16) & 95.44 \scriptsize (0.13) & 97.21 \scriptsize (0.13) & 90.92 \scriptsize (0.40) & 97.48 \scriptsize (0.06) & 97.39 \scriptsize (0.04) & \Aggr{0.473}{\hphantom{-}} \\
& & Ours &  98.21 \scriptsize (0.03) & 93.96 \scriptsize (0.17) & 95.72 \scriptsize (0.12) & 97.33 \scriptsize (0.10) & 91.63 \scriptsize (0.31) & 97.74 \scriptsize (0.09) & 97.52 \scriptsize (0.03) & \Aggr{0.745}{\bf \hphantom{-}} \\ \cmidrule{2-11}
& \multirow{2}{*}{Uni.} & Orig. &  98.04 \scriptsize (0.06) & 93.86 \scriptsize (0.18) & 95.66 \scriptsize (0.06) & 97.16 \scriptsize (0.07) & 91.12 \scriptsize (0.44) & 97.61 \scriptsize (0.07) & 97.35 \scriptsize (0.05) & \Aggr{0.541}{\hphantom{-}} \\
& & Ours &  98.11 \scriptsize (0.04) & 93.95 \scriptsize (0.18) & 95.72 \scriptsize (0.14) & 97.29 \scriptsize (0.11) & 91.51 \scriptsize (0.29) & 97.75 \scriptsize (0.09) & 97.52 \scriptsize (0.04) & \Aggr{0.712}{\hphantom{-}} \\ \bottomrule
\end{tabular}
}
\caption{Test accuracies for POS tagging including standard deviations over 10 random seeds and simple baselines from NTLK.}\label{tab:pos_appendix}
\end{table*}

\section{Additional Results}\label{app:results}

\begin{figure*}[ht]
\centering
Precision \\
\includegraphics[width=\textwidth]{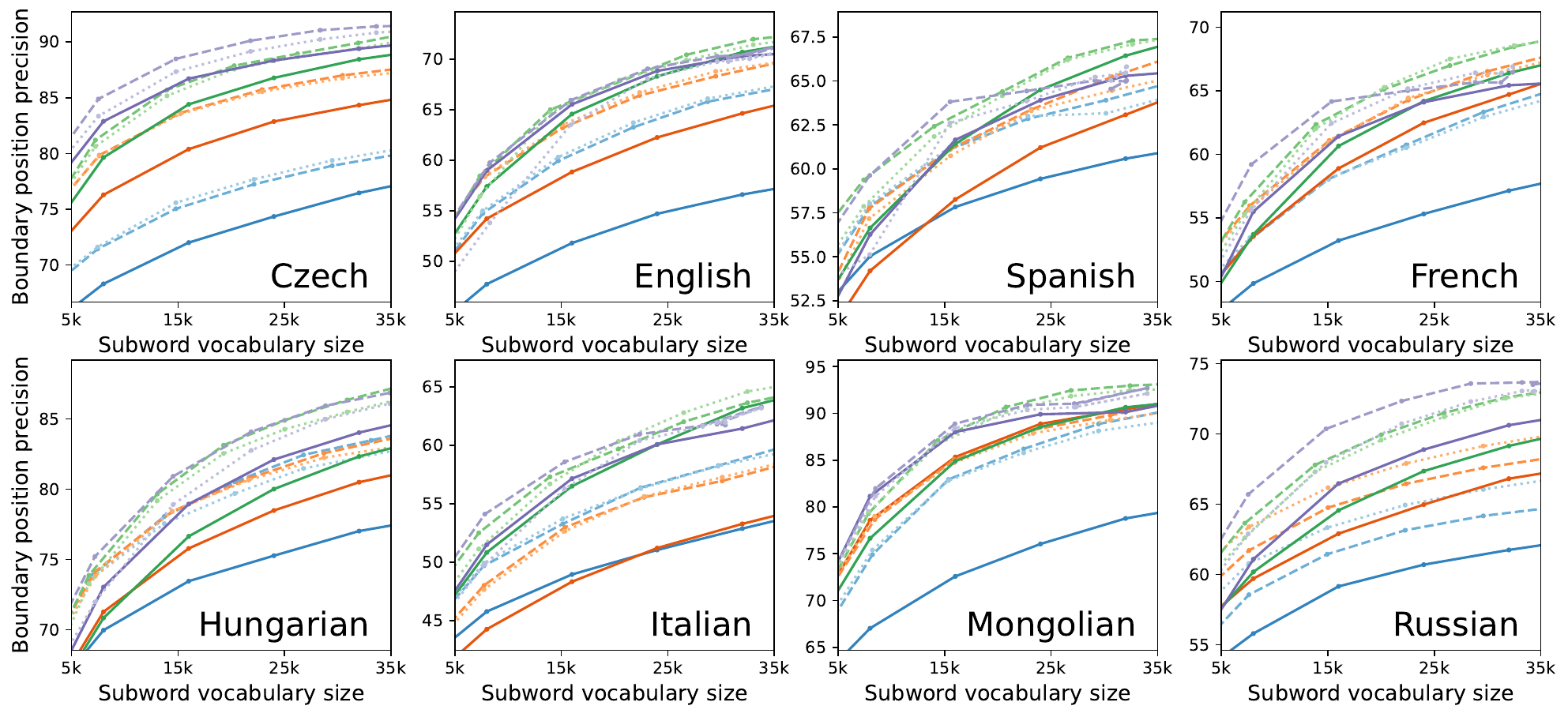} \\
Recall \\
\includegraphics[width=\textwidth]{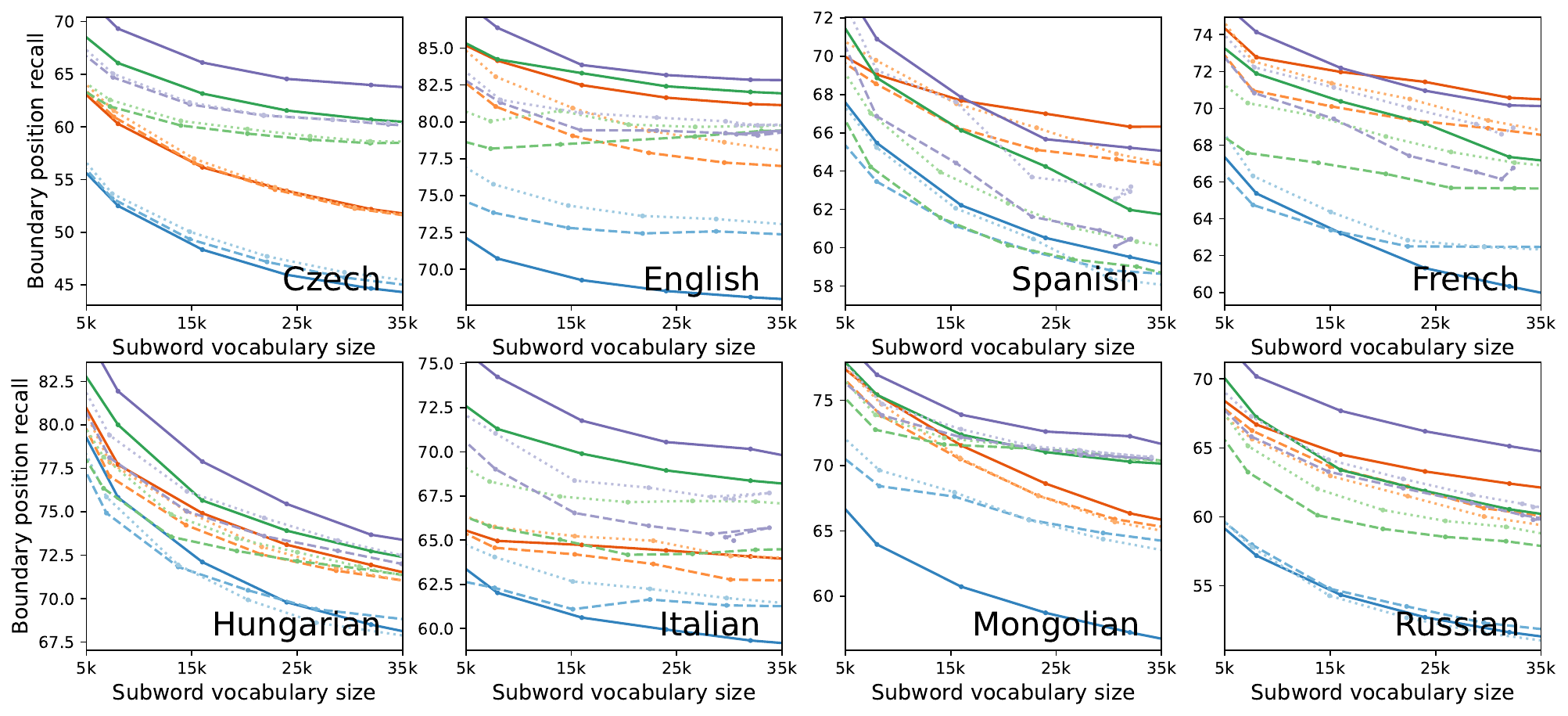} 
F1-Score \\
\includegraphics[width=\textwidth]{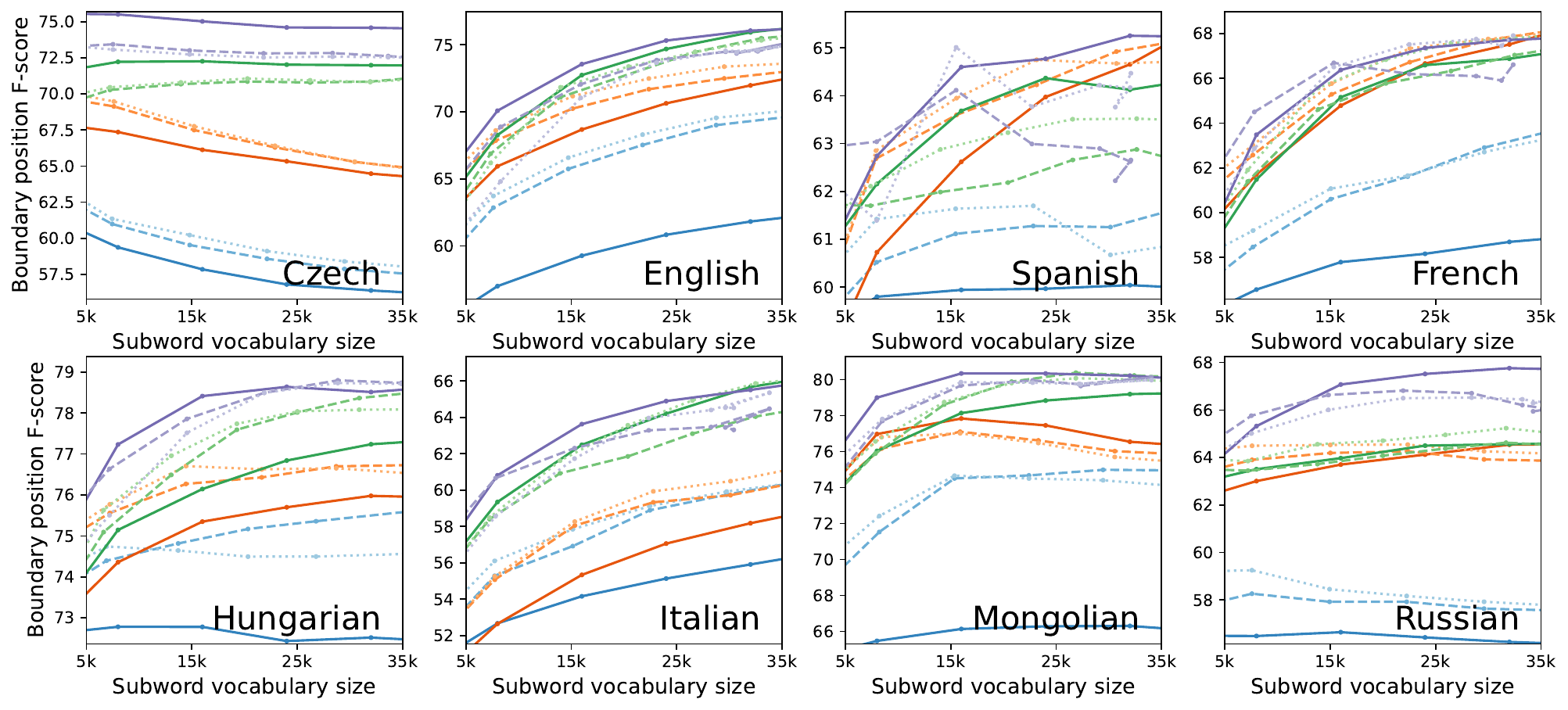}

\definecolor{plot1}{rgb}{0.19215686274509805, 0.5098039215686274, 0.7411764705882353}
\definecolor{plot2}{rgb}{0.9019607843137255, 0.3333333333333333, 0.050980392156862744}
\definecolor{plot3}{rgb}{0.19215686274509805, 0.6392156862745098, 0.32941176470588235}
\definecolor{plot4}{rgb}{0.4588235294117647, 0.4196078431372549, 0.6941176470588235}

\sf\footnotesize\centering
\textcolor{plot1}{$\bullet$} Word + BPE \quad \textcolor{plot2}{$\bullet$} Word + Unigram \quad \textcolor{plot3}{$\bullet$} Morfessor + BPE \quad \textcolor{plot4}{$\bullet$} Morfessor + Unigram \\
\tikz[baseline=-0.5ex]\draw [thick] (0,0) -- (0.5,0); Original \quad \tikz[baseline=-0.5ex]\draw [thick,dashed] (0,0) -- (0.5,0); Embedding-based inference \quad \tikz[baseline=-0.5ex]\draw [thick,dotted] (0,0) -- (.5,0); Bigram model

\caption{Boundary precision, recall and F1-score on the SIGMORPHON 2018 test set.}\label{fig:allIntrinsic}

\end{figure*}

Here, we present additional results: Precision, recall, and F$_1$ Score on the SIGMORPHON 2018 test set (Figure~\ref{fig:allIntrinsic}) and segmentations of randomly sampled words in Czech, English, and French (Table~\ref{tab:examples}). Table~\ref{tab:pos_appendix} contains more detailed results of POS tagging. Table~\ref{tab:bleu} contains aggregated BLEU scores for MT experiments, and Table~\ref{tab:chrf_all} contains individual chrF scores for the 18 language pairs.\punctfootnote{SacreBLEU signature: \texttt{BLEU|nrefs:1|case:mixed|\\eff:no|tok:13a|smooth:exp|version:2.0.0}}

\begin{table*}[ht]
\centering\footnotesize
    \begin{tabular}{p{.17\linewidth}p{.17\linewidth}p{.17\linewidth}p{.17\linewidth}p{.17\linewidth}}
         \toprule
        Word (Czech) & Gold segmentation & BPE & Unigram & Ours \\
         \midrule

        vykrášlit & vy\textcolor{Green}{\textvisiblespace}kráš\textcolor{Green}{\textvisiblespace}l\textcolor{Green}{\textvisiblespace}i\textcolor{Green}{\textvisiblespace}t & vy\textcolor{Green}{\textvisiblespace}krá\textcolor{Red}{\textvisiblespace}š\textcolor{Green}{\textvisiblespace}lit & vy\textcolor{Green}{\textvisiblespace}krá\textcolor{Red}{\textvisiblespace}š\textcolor{Green}{\textvisiblespace}lit & vy\textcolor{Green}{\textvisiblespace}krá\textcolor{Red}{\textvisiblespace}šl\textcolor{Green}{\textvisiblespace}it \\
fluorově & fluor\textcolor{Green}{\textvisiblespace}ov\textcolor{Green}{\textvisiblespace}ě & flu\textcolor{Red}{\textvisiblespace}or\textcolor{Green}{\textvisiblespace}ově & fl\textcolor{Red}{\textvisiblespace}u\textcolor{Red}{\textvisiblespace}or\textcolor{Green}{\textvisiblespace}ově & f\textcolor{Red}{\textvisiblespace}lu\textcolor{Red}{\textvisiblespace}or\textcolor{Green}{\textvisiblespace}ově \\
horách & hor\textcolor{Green}{\textvisiblespace}ách & horách & horách & horách \\
zkamenět & z\textcolor{Green}{\textvisiblespace}kamen\textcolor{Green}{\textvisiblespace}ě\textcolor{Green}{\textvisiblespace}t & z\textcolor{Green}{\textvisiblespace}kamen\textcolor{Green}{\textvisiblespace}ě\textcolor{Green}{\textvisiblespace}t & z\textcolor{Green}{\textvisiblespace}ka\textcolor{Red}{\textvisiblespace}me\textcolor{Red}{\textvisiblespace}ně\textcolor{Green}{\textvisiblespace}t & z\textcolor{Green}{\textvisiblespace}kamen\textcolor{Green}{\textvisiblespace}ě\textcolor{Green}{\textvisiblespace}t \\
akcií & akci\textcolor{Green}{\textvisiblespace}í & akcií & akcií & akcií \\
zdegenerovat & z\textcolor{Green}{\textvisiblespace}de\textcolor{Green}{\textvisiblespace}gener\textcolor{Green}{\textvisiblespace}ova\textcolor{Green}{\textvisiblespace}t & zde\textcolor{Green}{\textvisiblespace}gener\textcolor{Green}{\textvisiblespace}ovat & zde\textcolor{Green}{\textvisiblespace}gen\textcolor{Red}{\textvisiblespace}er\textcolor{Green}{\textvisiblespace}ovat & zde\textcolor{Green}{\textvisiblespace}gener\textcolor{Green}{\textvisiblespace}ovat \\
rezervy & re\textcolor{Green}{\textvisiblespace}zerv\textcolor{Green}{\textvisiblespace}y & rezervy & rezervy & rezervy \\
neměly & ne\textcolor{Green}{\textvisiblespace}m\textcolor{Green}{\textvisiblespace}ě\textcolor{Green}{\textvisiblespace}l\textcolor{Green}{\textvisiblespace}y & neměly & neměly & neměly \\
poplatků & po\textcolor{Green}{\textvisiblespace}plat\textcolor{Green}{\textvisiblespace}k\textcolor{Green}{\textvisiblespace}ů & poplatků & poplatků & poplatků \\
obnitkovat & ob\textcolor{Green}{\textvisiblespace}nit\textcolor{Green}{\textvisiblespace}k\textcolor{Green}{\textvisiblespace}ova\textcolor{Green}{\textvisiblespace}t & ob\textcolor{Green}{\textvisiblespace}ni\textcolor{Red}{\textvisiblespace}tk\textcolor{Green}{\textvisiblespace}ovat & ob\textcolor{Green}{\textvisiblespace}nit\textcolor{Green}{\textvisiblespace}kovat & ob\textcolor{Green}{\textvisiblespace}nit\textcolor{Green}{\textvisiblespace}kovat \\
znesnadňovat & z\textcolor{Green}{\textvisiblespace}ne\textcolor{Green}{\textvisiblespace}snad\textcolor{Green}{\textvisiblespace}ň\textcolor{Green}{\textvisiblespace}ova\textcolor{Green}{\textvisiblespace}t & zne\textcolor{Green}{\textvisiblespace}snad\textcolor{Green}{\textvisiblespace}ňovat & z\textcolor{Green}{\textvisiblespace}ne\textcolor{Green}{\textvisiblespace}snad\textcolor{Green}{\textvisiblespace}ňovat & zne\textcolor{Green}{\textvisiblespace}snad\textcolor{Green}{\textvisiblespace}ňovat \\
přesunovat & pře\textcolor{Green}{\textvisiblespace}sun\textcolor{Green}{\textvisiblespace}ova\textcolor{Green}{\textvisiblespace}t & přesu\textcolor{Red}{\textvisiblespace}novat & přesun\textcolor{Green}{\textvisiblespace}ovat & přesun\textcolor{Green}{\textvisiblespace}ovat \\
jednota & jedn\textcolor{Green}{\textvisiblespace}ot\textcolor{Green}{\textvisiblespace}a & jedno\textcolor{Red}{\textvisiblespace}ta & jedno\textcolor{Red}{\textvisiblespace}ta & jedno\textcolor{Red}{\textvisiblespace}ta \\
obklíčit & ob\textcolor{Green}{\textvisiblespace}klíč\textcolor{Green}{\textvisiblespace}i\textcolor{Green}{\textvisiblespace}t & ob\textcolor{Green}{\textvisiblespace}klí\textcolor{Red}{\textvisiblespace}čit & ob\textcolor{Green}{\textvisiblespace}klíč\textcolor{Green}{\textvisiblespace}it & ob\textcolor{Green}{\textvisiblespace}klíč\textcolor{Green}{\textvisiblespace}it \\
krysí & krys\textcolor{Green}{\textvisiblespace}í & kry\textcolor{Red}{\textvisiblespace}sí & krys\textcolor{Green}{\textvisiblespace}í & krys\textcolor{Green}{\textvisiblespace}í \\
premií & prem\textcolor{Green}{\textvisiblespace}i\textcolor{Green}{\textvisiblespace}í & premi\textcolor{Green}{\textvisiblespace}í & pre\textcolor{Red}{\textvisiblespace}mi\textcolor{Green}{\textvisiblespace}í & pre\textcolor{Red}{\textvisiblespace}mi\textcolor{Green}{\textvisiblespace}í \\
bříško & bříš\textcolor{Green}{\textvisiblespace}k\textcolor{Green}{\textvisiblespace}o & bří\textcolor{Red}{\textvisiblespace}ško & bříško & bříško \\
odpovídat & od\textcolor{Green}{\textvisiblespace}po\textcolor{Green}{\textvisiblespace}víd\textcolor{Green}{\textvisiblespace}a\textcolor{Green}{\textvisiblespace}t & odpovídat & odpovídat & odpovídat \\
zakuklit & za\textcolor{Green}{\textvisiblespace}kukl\textcolor{Green}{\textvisiblespace}i\textcolor{Green}{\textvisiblespace}t & za\textcolor{Green}{\textvisiblespace}ku\textcolor{Red}{\textvisiblespace}kli\textcolor{Green}{\textvisiblespace}t & za\textcolor{Green}{\textvisiblespace}ku\textcolor{Red}{\textvisiblespace}kli\textcolor{Green}{\textvisiblespace}t & za\textcolor{Green}{\textvisiblespace}kukl\textcolor{Green}{\textvisiblespace}it \\

        \\ \midrule
         Word (English) & Gold segmentation & BPE & Unigram & Ours \\
         \midrule
         
        macroclumps & macro\textcolor{Green}{\textvisiblespace}clump\textcolor{Green}{\textvisiblespace}s & macro\textcolor{Green}{\textvisiblespace}clum\textcolor{Red}{\textvisiblespace}ps & macro\textcolor{Green}{\textvisiblespace}cl\textcolor{Red}{\textvisiblespace}ump\textcolor{Green}{\textvisiblespace}s & macro\textcolor{Green}{\textvisiblespace}clump\textcolor{Green}{\textvisiblespace}s \\
gibbets & gibbet\textcolor{Green}{\textvisiblespace}s & gib\textcolor{Red}{\textvisiblespace}bets & gibb\textcolor{Red}{\textvisiblespace}ets & gibb\textcolor{Red}{\textvisiblespace}ets \\
phenoconverts & pheno\textcolor{Green}{\textvisiblespace}convert\textcolor{Green}{\textvisiblespace}s & phen\textcolor{Red}{\textvisiblespace}o\textcolor{Green}{\textvisiblespace}conver\textcolor{Red}{\textvisiblespace}ts & phe\textcolor{Red}{\textvisiblespace}no\textcolor{Green}{\textvisiblespace}con\textcolor{Red}{\textvisiblespace}vert\textcolor{Green}{\textvisiblespace}s & ph\textcolor{Red}{\textvisiblespace}eno\textcolor{Green}{\textvisiblespace}convert\textcolor{Green}{\textvisiblespace}s \\
ahura & ahura & a\textcolor{Red}{\textvisiblespace}hur\textcolor{Red}{\textvisiblespace}a & a\textcolor{Red}{\textvisiblespace}h\textcolor{Red}{\textvisiblespace}ura & ahu\textcolor{Red}{\textvisiblespace}ra \\
bimonopoles & bi\textcolor{Green}{\textvisiblespace}mono\textcolor{Green}{\textvisiblespace}pole\textcolor{Green}{\textvisiblespace}s & b\textcolor{Red}{\textvisiblespace}im\textcolor{Red}{\textvisiblespace}on\textcolor{Red}{\textvisiblespace}opol\textcolor{Red}{\textvisiblespace}es & bi\textcolor{Green}{\textvisiblespace}mon\textcolor{Red}{\textvisiblespace}o\textcolor{Green}{\textvisiblespace}pole\textcolor{Green}{\textvisiblespace}s & bi\textcolor{Green}{\textvisiblespace}mono\textcolor{Green}{\textvisiblespace}poles \\
nonwriter & non\textcolor{Green}{\textvisiblespace}write\textcolor{Green}{\textvisiblespace}r & non\textcolor{Green}{\textvisiblespace}writer & non\textcolor{Green}{\textvisiblespace}writer & non\textcolor{Green}{\textvisiblespace}writer \\
molelike & mole\textcolor{Green}{\textvisiblespace}like & mol\textcolor{Red}{\textvisiblespace}eli\textcolor{Red}{\textvisiblespace}ke & mole\textcolor{Green}{\textvisiblespace}like & mole\textcolor{Green}{\textvisiblespace}like \\
barnardsville & barnard\textcolor{Green}{\textvisiblespace}s\textcolor{Green}{\textvisiblespace}ville & bar\textcolor{Red}{\textvisiblespace}nar\textcolor{Red}{\textvisiblespace}d\textcolor{Green}{\textvisiblespace}sville & barnard\textcolor{Green}{\textvisiblespace}sville & barnard\textcolor{Green}{\textvisiblespace}sville \\
pogues & pogue\textcolor{Green}{\textvisiblespace}s & po\textcolor{Red}{\textvisiblespace}gues & po\textcolor{Red}{\textvisiblespace}gue\textcolor{Green}{\textvisiblespace}s & po\textcolor{Red}{\textvisiblespace}gu\textcolor{Red}{\textvisiblespace}es \\
infractors & infractor\textcolor{Green}{\textvisiblespace}s & infr\textcolor{Red}{\textvisiblespace}actors & in\textcolor{Red}{\textvisiblespace}fra\textcolor{Red}{\textvisiblespace}ctor\textcolor{Green}{\textvisiblespace}s & in\textcolor{Red}{\textvisiblespace}fr\textcolor{Red}{\textvisiblespace}actors \\
battlings & battling\textcolor{Green}{\textvisiblespace}s & batt\textcolor{Red}{\textvisiblespace}lings & battling\textcolor{Green}{\textvisiblespace}s & battling\textcolor{Green}{\textvisiblespace}s \\
larrup & larrup & lar\textcolor{Red}{\textvisiblespace}r\textcolor{Red}{\textvisiblespace}up & la\textcolor{Red}{\textvisiblespace}rr\textcolor{Red}{\textvisiblespace}up & lar\textcolor{Red}{\textvisiblespace}ru\textcolor{Red}{\textvisiblespace}p \\
detransformation & de\textcolor{Green}{\textvisiblespace}trans\textcolor{Green}{\textvisiblespace}form\textcolor{Green}{\textvisiblespace}ation & de\textcolor{Green}{\textvisiblespace}transformation & de\textcolor{Green}{\textvisiblespace}trans\textcolor{Green}{\textvisiblespace}form\textcolor{Green}{\textvisiblespace}ation & de\textcolor{Green}{\textvisiblespace}transform\textcolor{Green}{\textvisiblespace}ation \\
deexciting & de\textcolor{Green}{\textvisiblespace}excit\textcolor{Green}{\textvisiblespace}ing & de\textcolor{Green}{\textvisiblespace}exciting & de\textcolor{Green}{\textvisiblespace}ex\textcolor{Red}{\textvisiblespace}citing & de\textcolor{Green}{\textvisiblespace}exciting \\
kalasies & kalasie\textcolor{Green}{\textvisiblespace}s & kal\textcolor{Red}{\textvisiblespace}as\textcolor{Red}{\textvisiblespace}ies & kala\textcolor{Red}{\textvisiblespace}s\textcolor{Red}{\textvisiblespace}ies & kala\textcolor{Red}{\textvisiblespace}s\textcolor{Red}{\textvisiblespace}ies \\
canebrakes & cane\textcolor{Green}{\textvisiblespace}brake\textcolor{Green}{\textvisiblespace}s & can\textcolor{Red}{\textvisiblespace}e\textcolor{Green}{\textvisiblespace}brakes & can\textcolor{Red}{\textvisiblespace}e\textcolor{Green}{\textvisiblespace}bra\textcolor{Red}{\textvisiblespace}kes & ca\textcolor{Red}{\textvisiblespace}ne\textcolor{Green}{\textvisiblespace}brakes \\
eskimological & eskimo\textcolor{Green}{\textvisiblespace}log\textcolor{Green}{\textvisiblespace}ical & es\textcolor{Red}{\textvisiblespace}kim\textcolor{Red}{\textvisiblespace}ological & es\textcolor{Red}{\textvisiblespace}kim\textcolor{Red}{\textvisiblespace}ological & es\textcolor{Red}{\textvisiblespace}kim\textcolor{Red}{\textvisiblespace}ological \\
unmisleading & un\textcolor{Green}{\textvisiblespace}mis\textcolor{Green}{\textvisiblespace}lead\textcolor{Green}{\textvisiblespace}ing & un\textcolor{Green}{\textvisiblespace}misleading & un\textcolor{Green}{\textvisiblespace}mis\textcolor{Green}{\textvisiblespace}leading & un\textcolor{Green}{\textvisiblespace}misleading \\
neurofibromins & neuro\textcolor{Green}{\textvisiblespace}fib\textcolor{Green}{\textvisiblespace}r\textcolor{Green}{\textvisiblespace}om\textcolor{Green}{\textvisiblespace}in\textcolor{Green}{\textvisiblespace}s & neuro\textcolor{Green}{\textvisiblespace}fibro\textcolor{Red}{\textvisiblespace}mins & neuro\textcolor{Green}{\textvisiblespace}fi\textcolor{Red}{\textvisiblespace}bro\textcolor{Red}{\textvisiblespace}mins & neuro\textcolor{Green}{\textvisiblespace}fibro\textcolor{Red}{\textvisiblespace}mins \\

        \\ \midrule
         Word (French) & Gold segmentation & BPE & Unigram & Ours \\
         \midrule

        parassiens & parassien\textcolor{Green}{\textvisiblespace}s & par\textcolor{Red}{\textvisiblespace}assi\textcolor{Red}{\textvisiblespace}ens & par\textcolor{Red}{\textvisiblespace}assi\textcolor{Red}{\textvisiblespace}ens & pa\textcolor{Red}{\textvisiblespace}ras\textcolor{Red}{\textvisiblespace}siens \\
complaira & com\textcolor{Green}{\textvisiblespace}plair\textcolor{Green}{\textvisiblespace}a & compl\textcolor{Red}{\textvisiblespace}ai\textcolor{Red}{\textvisiblespace}ra & comp\textcolor{Red}{\textvisiblespace}la\textcolor{Red}{\textvisiblespace}ira & com\textcolor{Green}{\textvisiblespace}plaira \\
salindrois & salindr\textcolor{Green}{\textvisiblespace}ois & sal\textcolor{Red}{\textvisiblespace}in\textcolor{Red}{\textvisiblespace}dr\textcolor{Green}{\textvisiblespace}ois & sali\textcolor{Red}{\textvisiblespace}nd\textcolor{Red}{\textvisiblespace}rois & sali\textcolor{Red}{\textvisiblespace}nd\textcolor{Red}{\textvisiblespace}rois \\
nampontois & nampont\textcolor{Green}{\textvisiblespace}ois & nam\textcolor{Red}{\textvisiblespace}pon\textcolor{Red}{\textvisiblespace}tois & n\textcolor{Red}{\textvisiblespace}amp\textcolor{Red}{\textvisiblespace}ont\textcolor{Green}{\textvisiblespace}ois & nam\textcolor{Red}{\textvisiblespace}pont\textcolor{Green}{\textvisiblespace}ois \\
sédimentologique & sédimentologi\textcolor{Green}{\textvisiblespace}que & sé\textcolor{Red}{\textvisiblespace}di\textcolor{Red}{\textvisiblespace}ment\textcolor{Red}{\textvisiblespace}ologique & s\textcolor{Red}{\textvisiblespace}édi\textcolor{Red}{\textvisiblespace}ment\textcolor{Red}{\textvisiblespace}ologique & sé\textcolor{Red}{\textvisiblespace}dim\textcolor{Red}{\textvisiblespace}ent\textcolor{Red}{\textvisiblespace}ologique \\
esquivée & esquiv\textcolor{Green}{\textvisiblespace}é\textcolor{Green}{\textvisiblespace}e & esqui\textcolor{Red}{\textvisiblespace}vée & es\textcolor{Red}{\textvisiblespace}qui\textcolor{Red}{\textvisiblespace}vé\textcolor{Green}{\textvisiblespace}e & es\textcolor{Red}{\textvisiblespace}qu\textcolor{Red}{\textvisiblespace}iv\textcolor{Green}{\textvisiblespace}ée \\
flanc-garde & flanc\textcolor{Green}{\textvisiblespace}-\textcolor{Green}{\textvisiblespace}garde & fl\textcolor{Red}{\textvisiblespace}anc\textcolor{Green}{\textvisiblespace}-\textcolor{Green}{\textvisiblespace}garde & flanc\textcolor{Green}{\textvisiblespace}-\textcolor{Green}{\textvisiblespace}garde & flanc\textcolor{Green}{\textvisiblespace}-\textcolor{Green}{\textvisiblespace}garde \\
moyen & moyen & moyen & moyen & moyen \\
antigangs & anti\textcolor{Green}{\textvisiblespace}gang\textcolor{Green}{\textvisiblespace}s & anti\textcolor{Green}{\textvisiblespace}gangs & anti\textcolor{Green}{\textvisiblespace}g\textcolor{Red}{\textvisiblespace}ang\textcolor{Green}{\textvisiblespace}s & anti\textcolor{Green}{\textvisiblespace}gangs \\
forer & forer & for\textcolor{Red}{\textvisiblespace}er & for\textcolor{Red}{\textvisiblespace}er & fo\textcolor{Red}{\textvisiblespace}rer \\
captivités & captivité\textcolor{Green}{\textvisiblespace}s & capti\textcolor{Red}{\textvisiblespace}vités & captivité\textcolor{Green}{\textvisiblespace}s & captivité\textcolor{Green}{\textvisiblespace}s \\
dépolymérisés & dé\textcolor{Green}{\textvisiblespace}poly\textcolor{Green}{\textvisiblespace}m\textcolor{Green}{\textvisiblespace}é\textcolor{Green}{\textvisiblespace}r\textcolor{Green}{\textvisiblespace}is\textcolor{Green}{\textvisiblespace}é\textcolor{Green}{\textvisiblespace}s & dé\textcolor{Green}{\textvisiblespace}poly\textcolor{Green}{\textvisiblespace}m\textcolor{Green}{\textvisiblespace}ér\textcolor{Green}{\textvisiblespace}isés & dé\textcolor{Green}{\textvisiblespace}po\textcolor{Red}{\textvisiblespace}ly\textcolor{Green}{\textvisiblespace}mé\textcolor{Green}{\textvisiblespace}r\textcolor{Green}{\textvisiblespace}isés & dé\textcolor{Green}{\textvisiblespace}poly\textcolor{Green}{\textvisiblespace}mé\textcolor{Green}{\textvisiblespace}ris\textcolor{Green}{\textvisiblespace}és \\
prévoiriez & pré\textcolor{Green}{\textvisiblespace}voir\textcolor{Green}{\textvisiblespace}iez & pré\textcolor{Green}{\textvisiblespace}voi\textcolor{Red}{\textvisiblespace}riez & prévoir\textcolor{Green}{\textvisiblespace}iez & prévoir\textcolor{Green}{\textvisiblespace}iez \\
déracinerais & dé\textcolor{Green}{\textvisiblespace}racine\textcolor{Green}{\textvisiblespace}r\textcolor{Green}{\textvisiblespace}ais & dé\textcolor{Green}{\textvisiblespace}rac\textcolor{Red}{\textvisiblespace}in\textcolor{Red}{\textvisiblespace}erais & d\textcolor{Red}{\textvisiblespace}éra\textcolor{Red}{\textvisiblespace}cine\textcolor{Green}{\textvisiblespace}rais & dé\textcolor{Green}{\textvisiblespace}racine\textcolor{Green}{\textvisiblespace}rais \\
corécipiendaire & co\textcolor{Green}{\textvisiblespace}récipiendaire & coré\textcolor{Red}{\textvisiblespace}ci\textcolor{Red}{\textvisiblespace}pi\textcolor{Red}{\textvisiblespace}end\textcolor{Red}{\textvisiblespace}aire & cor\textcolor{Red}{\textvisiblespace}é\textcolor{Red}{\textvisiblespace}ci\textcolor{Red}{\textvisiblespace}pi\textcolor{Red}{\textvisiblespace}end\textcolor{Red}{\textvisiblespace}aire & co\textcolor{Green}{\textvisiblespace}ré\textcolor{Red}{\textvisiblespace}cip\textcolor{Red}{\textvisiblespace}ien\textcolor{Red}{\textvisiblespace}da\textcolor{Red}{\textvisiblespace}ire \\
crustacyanines & crustacyanine\textcolor{Green}{\textvisiblespace}s & cru\textcolor{Red}{\textvisiblespace}st\textcolor{Red}{\textvisiblespace}ac\textcolor{Red}{\textvisiblespace}yan\textcolor{Red}{\textvisiblespace}ines & crus\textcolor{Red}{\textvisiblespace}t\textcolor{Red}{\textvisiblespace}ac\textcolor{Red}{\textvisiblespace}yan\textcolor{Red}{\textvisiblespace}ines & crus\textcolor{Red}{\textvisiblespace}tac\textcolor{Red}{\textvisiblespace}yan\textcolor{Red}{\textvisiblespace}ines \\
chambardés & chambard\textcolor{Green}{\textvisiblespace}é\textcolor{Green}{\textvisiblespace}s & cham\textcolor{Red}{\textvisiblespace}bar\textcolor{Red}{\textvisiblespace}dés & chamb\textcolor{Red}{\textvisiblespace}ard\textcolor{Green}{\textvisiblespace}és & cham\textcolor{Red}{\textvisiblespace}bard\textcolor{Green}{\textvisiblespace}és \\
joyeusain & joyeus\textcolor{Green}{\textvisiblespace}ain & joy\textcolor{Red}{\textvisiblespace}eu\textcolor{Red}{\textvisiblespace}sain & joy\textcolor{Red}{\textvisiblespace}e\textcolor{Red}{\textvisiblespace}us\textcolor{Green}{\textvisiblespace}ain & jo\textcolor{Red}{\textvisiblespace}ye\textcolor{Red}{\textvisiblespace}usain \\
influions & influ\textcolor{Green}{\textvisiblespace}ions & influ\textcolor{Green}{\textvisiblespace}ions & in\textcolor{Red}{\textvisiblespace}flu\textcolor{Green}{\textvisiblespace}ions & inf\textcolor{Red}{\textvisiblespace}lu\textcolor{Green}{\textvisiblespace}ions \\

        \\ \bottomrule
    \end{tabular}

    \caption{Example segmentations from the SIGMORPHON 2018 Czech, English, and French test sets. \textcolor{Green}{Green} space symbols denote morphologically valid splits, and the \textcolor{Red}{red} space symbols denote splits inside morphemes.}\label{tab:examples}
\end{table*}

\begin{table*}[ht]

\footnotesize\centering
\begin{minipage}{.33\textwidth}\scalebox{0.75}{%
\newcommand{\ChrfaraXeng}[2]{\gradientcell{#1}{38.090379444419646}{41.77770613126815}{Cyan}{Yellow}{60}{#2}}
\begin{tabular}{lll ccc@{\hskip 18pt}c}
\toprule
\multicolumn{3}{l}{\multirow{2}{*}{\Large ara-eng}} & \multicolumn{3}{c}{Vocabulary\hspace*{15pt}} & \multirow{2}{*}{Avg.} \\ \cmidrule(lr{18pt}){4-6}
& & & 4k & 8k & 16k & \\ \midrule

\multirow{4}{*}{\rotatebox[origin=c]{90}{Word-like}} & \multirow{2}{*}{BPE} & Orig. & \ChrfaraXeng{38.8}{\hphantom{-}} & \ChrfaraXeng{39.4}{\hphantom{-}} & \ChrfaraXeng{39.9}{\hphantom{-}} & \ChrfaraXeng{39.4}{\hphantom{-}} \\
& & Ours & \ChrfaraXeng{38.6}{\hphantom{-}} & \ChrfaraXeng{39.8}{\hphantom{-}} & \ChrfaraXeng{40.3}{\hphantom{-}} & \ChrfaraXeng{39.6}{\hphantom{-}} \\ \cmidrule{2-7}
& \multirow{2}{*}{Uni.} & Orig. & \ChrfaraXeng{40.1}{\hphantom{-}} & \ChrfaraXeng{40.6}{\hphantom{-}} & \ChrfaraXeng{41.3}{\hphantom{-}} & \ChrfaraXeng{40.7}{\hphantom{-}} \\
& & Ours & \ChrfaraXeng{39.4}{\hphantom{-}} & \ChrfaraXeng{40.6}{\hphantom{-}} & \ChrfaraXeng{41.0}{\hphantom{-}} & \ChrfaraXeng{40.4}{\hphantom{-}} \\ \midrule
\multirow{4}{*}{\rotatebox[origin=c]{90}{Morfessor}} & \multirow{2}{*}{BPE} & Orig. & \ChrfaraXeng{39.9}{\hphantom{-}} & \ChrfaraXeng{40.1}{\hphantom{-}} & \ChrfaraXeng{40.7}{\hphantom{-}} & \ChrfaraXeng{40.2}{\hphantom{-}} \\
& & Ours & \ChrfaraXeng{39.4}{\hphantom{-}} & \ChrfaraXeng{40.2}{\hphantom{-}} & \ChrfaraXeng{41.0}{\hphantom{-}} & \ChrfaraXeng{40.2}{\hphantom{-}} \\ \cmidrule{2-7}
& \multirow{2}{*}{Uni.} & Orig. & \ChrfaraXeng{39.5}{\hphantom{-}} & \ChrfaraXeng{40.5}{\hphantom{-}} & \ChrfaraXeng{40.0}{\hphantom{-}} & \ChrfaraXeng{40.0}{\hphantom{-}} \\
& & Ours & \ChrfaraXeng{40.0}{\hphantom{-}} & \ChrfaraXeng{41.3}{\hphantom{-}} & \ChrfaraXeng{40.1}{\hphantom{-}} & \ChrfaraXeng{40.5}{\hphantom{-}} \\ \bottomrule
\end{tabular}
}\end{minipage}
\begin{minipage}{.33\textwidth}\scalebox{0.75}{%
\newcommand{\ChrfengXara}[2]{\gradientcell{#1}{32.843513672700496}{35.70782157253844}{Cyan}{Yellow}{60}{#2}}
\begin{tabular}{lll ccc@{\hskip 18pt}c}
\toprule
\multicolumn{3}{l}{\multirow{2}{*}{\Large eng-ara}} & \multicolumn{3}{c}{Vocabulary\hspace*{15pt}} & \multirow{2}{*}{Avg.} \\ \cmidrule(lr{18pt}){4-6}
& & & 4k & 8k & 16k & \\ \midrule

\multirow{4}{*}{\rotatebox[origin=c]{90}{Word-like}} & \multirow{2}{*}{BPE} & Orig. & \ChrfengXara{33.4}{\hphantom{-}} & \ChrfengXara{34.2}{\hphantom{-}} & \ChrfengXara{34.7}{\hphantom{-}} & \ChrfengXara{34.1}{\hphantom{-}} \\
& & Ours & \ChrfengXara{33.4}{\hphantom{-}} & \ChrfengXara{34.0}{\hphantom{-}} & \ChrfengXara{34.4}{\hphantom{-}} & \ChrfengXara{33.9}{\hphantom{-}} \\ \cmidrule{2-7}
& \multirow{2}{*}{Uni.} & Orig. & \ChrfengXara{33.9}{\hphantom{-}} & \ChrfengXara{35.2}{\hphantom{-}} & \ChrfengXara{34.7}{\hphantom{-}} & \ChrfengXara{34.6}{\hphantom{-}} \\
& & Ours & \ChrfengXara{33.9}{\hphantom{-}} & \ChrfengXara{34.4}{\hphantom{-}} & \ChrfengXara{34.9}{\hphantom{-}} & \ChrfengXara{34.4}{\hphantom{-}} \\ \midrule
\multirow{4}{*}{\rotatebox[origin=c]{90}{Morfessor}} & \multirow{2}{*}{BPE} & Orig. & \ChrfengXara{33.6}{\hphantom{-}} & \ChrfengXara{34.5}{\hphantom{-}} & \ChrfengXara{34.5}{\hphantom{-}} & \ChrfengXara{34.2}{\hphantom{-}} \\
& & Ours & \ChrfengXara{34.1}{\hphantom{-}} & \ChrfengXara{35.0}{\hphantom{-}} & \ChrfengXara{35.1}{\hphantom{-}} & \ChrfengXara{34.7}{\hphantom{-}} \\ \cmidrule{2-7}
& \multirow{2}{*}{Uni.} & Orig. & \ChrfengXara{33.3}{\hphantom{-}} & \ChrfengXara{34.3}{\hphantom{-}} & \ChrfengXara{34.5}{\hphantom{-}} & \ChrfengXara{34.0}{\hphantom{-}} \\
& & Ours & \ChrfengXara{33.9}{\hphantom{-}} & \ChrfengXara{35.0}{\hphantom{-}} & \ChrfengXara{34.3}{\hphantom{-}} & \ChrfengXara{34.4}{\hphantom{-}} \\ \bottomrule
\end{tabular}
}\end{minipage}
\begin{minipage}{.33\textwidth}\scalebox{0.75}{%
\newcommand{\ChrfdeuXeng}[2]{\gradientcell{#1}{41.134491209966725}{45.289648774653536}{Cyan}{Yellow}{60}{#2}}
\begin{tabular}{lll ccc@{\hskip 18pt}c}
\toprule
\multicolumn{3}{l}{\multirow{2}{*}{\Large deu-eng}} & \multicolumn{3}{c}{Vocabulary\hspace*{15pt}} & \multirow{2}{*}{Avg.} \\ \cmidrule(lr{18pt}){4-6}
& & & 4k & 8k & 16k & \\ \midrule

\multirow{4}{*}{\rotatebox[origin=c]{90}{Word-like}} & \multirow{2}{*}{BPE} & Orig. & \ChrfdeuXeng{43.6}{\hphantom{-}} & \ChrfdeuXeng{43.8}{\hphantom{-}} & \ChrfdeuXeng{43.9}{\hphantom{-}} & \ChrfdeuXeng{43.8}{\hphantom{-}} \\
& & Ours & \ChrfdeuXeng{43.2}{\hphantom{-}} & \ChrfdeuXeng{43.8}{\hphantom{-}} & \ChrfdeuXeng{43.9}{\hphantom{-}} & \ChrfdeuXeng{43.6}{\hphantom{-}} \\ \cmidrule{2-7}
& \multirow{2}{*}{Uni.} & Orig. & \ChrfdeuXeng{42.9}{\hphantom{-}} & \ChrfdeuXeng{44.3}{\hphantom{-}} & \ChrfdeuXeng{44.8}{\hphantom{-}} & \ChrfdeuXeng{44.0}{\hphantom{-}} \\
& & Ours & \ChrfdeuXeng{43.7}{\hphantom{-}} & \ChrfdeuXeng{44.1}{\hphantom{-}} & \ChrfdeuXeng{44.6}{\hphantom{-}} & \ChrfdeuXeng{44.1}{\hphantom{-}} \\ \midrule
\multirow{4}{*}{\rotatebox[origin=c]{90}{Morfessor}} & \multirow{2}{*}{BPE} & Orig. & \ChrfdeuXeng{42.6}{\hphantom{-}} & \ChrfdeuXeng{42.3}{\hphantom{-}} & \ChrfdeuXeng{42.1}{\hphantom{-}} & \ChrfdeuXeng{42.3}{\hphantom{-}} \\
& & Ours & \ChrfdeuXeng{43.1}{\hphantom{-}} & \ChrfdeuXeng{43.2}{\hphantom{-}} & \ChrfdeuXeng{43.5}{\hphantom{-}} & \ChrfdeuXeng{43.2}{\hphantom{-}} \\ \cmidrule{2-7}
& \multirow{2}{*}{Uni.} & Orig. & \ChrfdeuXeng{41.9}{\hphantom{-}} & \ChrfdeuXeng{42.0}{\hphantom{-}} & \ChrfdeuXeng{41.6}{\hphantom{-}} & \ChrfdeuXeng{41.9}{\hphantom{-}} \\
& & Ours & \ChrfdeuXeng{43.2}{\hphantom{-}} & \ChrfdeuXeng{42.9}{\hphantom{-}} & \ChrfdeuXeng{43.5}{\hphantom{-}} & \ChrfdeuXeng{43.2}{\hphantom{-}} \\ \bottomrule
\end{tabular}
}\end{minipage}
\\[1.8ex]
\begin{minipage}{.33\textwidth}\scalebox{0.75}{%
\newcommand{\ChrfengXdeu}[2]{\gradientcell{#1}{41.67850266712695}{46.655385924794686}{Cyan}{Yellow}{60}{#2}}
\begin{tabular}{lll ccc@{\hskip 18pt}c}
\toprule
\multicolumn{3}{l}{\multirow{2}{*}{\Large eng-deu}} & \multicolumn{3}{c}{Vocabulary\hspace*{15pt}} & \multirow{2}{*}{Avg.} \\ \cmidrule(lr{18pt}){4-6}
& & & 4k & 8k & 16k & \\ \midrule

\multirow{4}{*}{\rotatebox[origin=c]{90}{Word-like}} & \multirow{2}{*}{BPE} & Orig. & \ChrfengXdeu{44.8}{\hphantom{-}} & \ChrfengXdeu{45.4}{\hphantom{-}} & \ChrfengXdeu{45.0}{\hphantom{-}} & \ChrfengXdeu{45.1}{\hphantom{-}} \\
& & Ours & \ChrfengXdeu{44.2}{\hphantom{-}} & \ChrfengXdeu{45.5}{\hphantom{-}} & \ChrfengXdeu{45.0}{\hphantom{-}} & \ChrfengXdeu{44.9}{\hphantom{-}} \\ \cmidrule{2-7}
& \multirow{2}{*}{Uni.} & Orig. & \ChrfengXdeu{44.4}{\hphantom{-}} & \ChrfengXdeu{45.6}{\hphantom{-}} & \ChrfengXdeu{45.6}{\hphantom{-}} & \ChrfengXdeu{45.2}{\hphantom{-}} \\
& & Ours & \ChrfengXdeu{44.6}{\hphantom{-}} & \ChrfengXdeu{44.6}{\hphantom{-}} & \ChrfengXdeu{46.2}{\hphantom{-}} & \ChrfengXdeu{45.1}{\hphantom{-}} \\ \midrule
\multirow{4}{*}{\rotatebox[origin=c]{90}{Morfessor}} & \multirow{2}{*}{BPE} & Orig. & \ChrfengXdeu{42.2}{\hphantom{-}} & \ChrfengXdeu{44.1}{\hphantom{-}} & \ChrfengXdeu{45.0}{\hphantom{-}} & \ChrfengXdeu{43.8}{\hphantom{-}} \\
& & Ours & \ChrfengXdeu{44.3}{\hphantom{-}} & \ChrfengXdeu{44.8}{\hphantom{-}} & \ChrfengXdeu{45.0}{\hphantom{-}} & \ChrfengXdeu{44.7}{\hphantom{-}} \\ \cmidrule{2-7}
& \multirow{2}{*}{Uni.} & Orig. & \ChrfengXdeu{43.2}{\hphantom{-}} & \ChrfengXdeu{43.5}{\hphantom{-}} & \ChrfengXdeu{43.8}{\hphantom{-}} & \ChrfengXdeu{43.5}{\hphantom{-}} \\
& & Ours & \ChrfengXdeu{44.6}{\hphantom{-}} & \ChrfengXdeu{45.2}{\hphantom{-}} & \ChrfengXdeu{44.3}{\hphantom{-}} & \ChrfengXdeu{44.7}{\hphantom{-}} \\ \bottomrule
\end{tabular}
}\end{minipage}
\begin{minipage}{.33\textwidth}\scalebox{0.75}{%
\newcommand{\ChrffraXeng}[2]{\gradientcell{#1}{48.00630103603021}{52.124466198534314}{Cyan}{Yellow}{60}{#2}}
\begin{tabular}{lll ccc@{\hskip 18pt}c}
\toprule
\multicolumn{3}{l}{\multirow{2}{*}{\Large fra-eng}} & \multicolumn{3}{c}{Vocabulary\hspace*{15pt}} & \multirow{2}{*}{Avg.} \\ \cmidrule(lr{18pt}){4-6}
& & & 4k & 8k & 16k & \\ \midrule

\multirow{4}{*}{\rotatebox[origin=c]{90}{Word-like}} & \multirow{2}{*}{BPE} & Orig. & \ChrffraXeng{50.3}{\hphantom{-}} & \ChrffraXeng{51.3}{\hphantom{-}} & \ChrffraXeng{51.1}{\hphantom{-}} & \ChrffraXeng{50.9}{\hphantom{-}} \\
& & Ours & \ChrffraXeng{50.3}{\hphantom{-}} & \ChrffraXeng{50.9}{\hphantom{-}} & \ChrffraXeng{51.6}{\hphantom{-}} & \ChrffraXeng{50.9}{\hphantom{-}} \\ \cmidrule{2-7}
& \multirow{2}{*}{Uni.} & Orig. & \ChrffraXeng{50.1}{\hphantom{-}} & \ChrffraXeng{51.1}{\hphantom{-}} & \ChrffraXeng{51.6}{\hphantom{-}} & \ChrffraXeng{50.9}{\hphantom{-}} \\
& & Ours & \ChrffraXeng{50.0}{\hphantom{-}} & \ChrffraXeng{51.6}{\hphantom{-}} & \ChrffraXeng{49.3}{\hphantom{-}} & \ChrffraXeng{50.3}{\hphantom{-}} \\ \midrule
\multirow{4}{*}{\rotatebox[origin=c]{90}{Morfessor}} & \multirow{2}{*}{BPE} & Orig. & \ChrffraXeng{48.9}{\hphantom{-}} & \ChrffraXeng{48.8}{\hphantom{-}} & \ChrffraXeng{48.9}{\hphantom{-}} & \ChrffraXeng{48.9}{\hphantom{-}} \\
& & Ours & \ChrffraXeng{50.5}{\hphantom{-}} & \ChrffraXeng{50.3}{\hphantom{-}} & \ChrffraXeng{50.2}{\hphantom{-}} & \ChrffraXeng{50.3}{\hphantom{-}} \\ \cmidrule{2-7}
& \multirow{2}{*}{Uni.} & Orig. & \ChrffraXeng{48.6}{\hphantom{-}} & \ChrffraXeng{49.0}{\hphantom{-}} & \ChrffraXeng{48.5}{\hphantom{-}} & \ChrffraXeng{48.7}{\hphantom{-}} \\
& & Ours & \ChrffraXeng{49.4}{\hphantom{-}} & \ChrffraXeng{50.3}{\hphantom{-}} & \ChrffraXeng{49.5}{\hphantom{-}} & \ChrffraXeng{49.7}{\hphantom{-}} \\ \bottomrule
\end{tabular}
}\end{minipage}
\begin{minipage}{.33\textwidth}\scalebox{0.75}{%
\newcommand{\ChrfengXfra}[2]{\gradientcell{#1}{50.21026955761025}{53.923607832694685}{Cyan}{Yellow}{60}{#2}}
\begin{tabular}{lll ccc@{\hskip 18pt}c}
\toprule
\multicolumn{3}{l}{\multirow{2}{*}{\Large eng-fra}} & \multicolumn{3}{c}{Vocabulary\hspace*{15pt}} & \multirow{2}{*}{Avg.} \\ \cmidrule(lr{18pt}){4-6}
& & & 4k & 8k & 16k & \\ \midrule

\multirow{4}{*}{\rotatebox[origin=c]{90}{Word-like}} & \multirow{2}{*}{BPE} & Orig. & \ChrfengXfra{52.1}{\hphantom{-}} & \ChrfengXfra{52.9}{\hphantom{-}} & \ChrfengXfra{53.2}{\hphantom{-}} & \ChrfengXfra{52.7}{\hphantom{-}} \\
& & Ours & \ChrfengXfra{52.5}{\hphantom{-}} & \ChrfengXfra{52.3}{\hphantom{-}} & \ChrfengXfra{52.9}{\hphantom{-}} & \ChrfengXfra{52.6}{\hphantom{-}} \\ \cmidrule{2-7}
& \multirow{2}{*}{Uni.} & Orig. & \ChrfengXfra{52.1}{\hphantom{-}} & \ChrfengXfra{53.4}{\hphantom{-}} & \ChrfengXfra{53.4}{\hphantom{-}} & \ChrfengXfra{53.0}{\hphantom{-}} \\
& & Ours & \ChrfengXfra{51.6}{\hphantom{-}} & \ChrfengXfra{53.0}{\hphantom{-}} & \ChrfengXfra{51.0}{\hphantom{-}} & \ChrfengXfra{51.9}{\hphantom{-}} \\ \midrule
\multirow{4}{*}{\rotatebox[origin=c]{90}{Morfessor}} & \multirow{2}{*}{BPE} & Orig. & \ChrfengXfra{50.7}{\hphantom{-}} & \ChrfengXfra{51.5}{\hphantom{-}} & \ChrfengXfra{51.8}{\hphantom{-}} & \ChrfengXfra{51.3}{\hphantom{-}} \\
& & Ours & \ChrfengXfra{51.8}{\hphantom{-}} & \ChrfengXfra{52.0}{\hphantom{-}} & \ChrfengXfra{53.1}{\hphantom{-}} & \ChrfengXfra{52.3}{\hphantom{-}} \\ \cmidrule{2-7}
& \multirow{2}{*}{Uni.} & Orig. & \ChrfengXfra{51.0}{\hphantom{-}} & \ChrfengXfra{51.7}{\hphantom{-}} & \ChrfengXfra{51.6}{\hphantom{-}} & \ChrfengXfra{51.4}{\hphantom{-}} \\
& & Ours & \ChrfengXfra{52.2}{\hphantom{-}} & \ChrfengXfra{52.8}{\hphantom{-}} & \ChrfengXfra{52.3}{\hphantom{-}} & \ChrfengXfra{52.4}{\hphantom{-}} \\ \bottomrule
\end{tabular}
}\end{minipage}
\\[1.8ex]
\begin{minipage}{.33\textwidth}\scalebox{0.75}{%
\newcommand{\ChrfnldXeng}[2]{\gradientcell{#1}{46.136743815371}{49.1131922177645}{Cyan}{Yellow}{60}{#2}}
\begin{tabular}{lll ccc@{\hskip 18pt}c}
\toprule
\multicolumn{3}{l}{\multirow{2}{*}{\Large nld-eng}} & \multicolumn{3}{c}{Vocabulary\hspace*{15pt}} & \multirow{2}{*}{Avg.} \\ \cmidrule(lr{18pt}){4-6}
& & & 4k & 8k & 16k & \\ \midrule

\multirow{4}{*}{\rotatebox[origin=c]{90}{Word-like}} & \multirow{2}{*}{BPE} & Orig. & \ChrfnldXeng{47.8}{\hphantom{-}} & \ChrfnldXeng{48.6}{\hphantom{-}} & \ChrfnldXeng{48.4}{\hphantom{-}} & \ChrfnldXeng{48.3}{\hphantom{-}} \\
& & Ours & \ChrfnldXeng{48.2}{\hphantom{-}} & \ChrfnldXeng{47.4}{\hphantom{-}} & \ChrfnldXeng{48.1}{\hphantom{-}} & \ChrfnldXeng{47.9}{\hphantom{-}} \\ \cmidrule{2-7}
& \multirow{2}{*}{Uni.} & Orig. & \ChrfnldXeng{48.4}{\hphantom{-}} & \ChrfnldXeng{48.3}{\hphantom{-}} & \ChrfnldXeng{48.2}{\hphantom{-}} & \ChrfnldXeng{48.3}{\hphantom{-}} \\
& & Ours & \ChrfnldXeng{47.7}{\hphantom{-}} & \ChrfnldXeng{48.6}{\hphantom{-}} & \ChrfnldXeng{48.4}{\hphantom{-}} & \ChrfnldXeng{48.3}{\hphantom{-}} \\ \midrule
\multirow{4}{*}{\rotatebox[origin=c]{90}{Morfessor}} & \multirow{2}{*}{BPE} & Orig. & \ChrfnldXeng{47.5}{\hphantom{-}} & \ChrfnldXeng{46.6}{\hphantom{-}} & \ChrfnldXeng{47.5}{\hphantom{-}} & \ChrfnldXeng{47.2}{\hphantom{-}} \\
& & Ours & \ChrfnldXeng{47.4}{\hphantom{-}} & \ChrfnldXeng{47.7}{\hphantom{-}} & \ChrfnldXeng{47.5}{\hphantom{-}} & \ChrfnldXeng{47.6}{\hphantom{-}} \\ \cmidrule{2-7}
& \multirow{2}{*}{Uni.} & Orig. & \ChrfnldXeng{46.8}{\hphantom{-}} & \ChrfnldXeng{47.1}{\hphantom{-}} & \ChrfnldXeng{46.7}{\hphantom{-}} & \ChrfnldXeng{46.8}{\hphantom{-}} \\
& & Ours & \ChrfnldXeng{47.5}{\hphantom{-}} & \ChrfnldXeng{47.2}{\hphantom{-}} & \ChrfnldXeng{47.2}{\hphantom{-}} & \ChrfnldXeng{47.3}{\hphantom{-}} \\ \bottomrule
\end{tabular}
}\end{minipage}
\begin{minipage}{.33\textwidth}\scalebox{0.75}{%
\newcommand{\ChrfengXnld}[2]{\gradientcell{#1}{45.0246538909505}{48.61178865901453}{Cyan}{Yellow}{60}{#2}}
\begin{tabular}{lll ccc@{\hskip 18pt}c}
\toprule
\multicolumn{3}{l}{\multirow{2}{*}{\Large eng-nld}} & \multicolumn{3}{c}{Vocabulary\hspace*{15pt}} & \multirow{2}{*}{Avg.} \\ \cmidrule(lr{18pt}){4-6}
& & & 4k & 8k & 16k & \\ \midrule

\multirow{4}{*}{\rotatebox[origin=c]{90}{Word-like}} & \multirow{2}{*}{BPE} & Orig. & \ChrfengXnld{47.4}{\hphantom{-}} & \ChrfengXnld{47.9}{\hphantom{-}} & \ChrfengXnld{47.6}{\hphantom{-}} & \ChrfengXnld{47.6}{\hphantom{-}} \\
& & Ours & \ChrfengXnld{45.9}{\hphantom{-}} & \ChrfengXnld{48.0}{\hphantom{-}} & \ChrfengXnld{47.7}{\hphantom{-}} & \ChrfengXnld{47.2}{\hphantom{-}} \\ \cmidrule{2-7}
& \multirow{2}{*}{Uni.} & Orig. & \ChrfengXnld{46.3}{\hphantom{-}} & \ChrfengXnld{48.1}{\hphantom{-}} & \ChrfengXnld{47.4}{\hphantom{-}} & \ChrfengXnld{47.3}{\hphantom{-}} \\
& & Ours & \ChrfengXnld{46.3}{\hphantom{-}} & \ChrfengXnld{47.3}{\hphantom{-}} & \ChrfengXnld{48.1}{\hphantom{-}} & \ChrfengXnld{47.2}{\hphantom{-}} \\ \midrule
\multirow{4}{*}{\rotatebox[origin=c]{90}{Morfessor}} & \multirow{2}{*}{BPE} & Orig. & \ChrfengXnld{45.7}{\hphantom{-}} & \ChrfengXnld{45.5}{\hphantom{-}} & \ChrfengXnld{46.6}{\hphantom{-}} & \ChrfengXnld{45.9}{\hphantom{-}} \\
& & Ours & \ChrfengXnld{46.7}{\hphantom{-}} & \ChrfengXnld{46.8}{\hphantom{-}} & \ChrfengXnld{47.7}{\hphantom{-}} & \ChrfengXnld{47.1}{\hphantom{-}} \\ \cmidrule{2-7}
& \multirow{2}{*}{Uni.} & Orig. & \ChrfengXnld{46.1}{\hphantom{-}} & \ChrfengXnld{46.0}{\hphantom{-}} & \ChrfengXnld{46.1}{\hphantom{-}} & \ChrfengXnld{46.1}{\hphantom{-}} \\
& & Ours & \ChrfengXnld{46.2}{\hphantom{-}} & \ChrfengXnld{47.4}{\hphantom{-}} & \ChrfengXnld{46.1}{\hphantom{-}} & \ChrfengXnld{46.6}{\hphantom{-}} \\ \bottomrule
\end{tabular}
}\end{minipage}
\begin{minipage}{.33\textwidth}\scalebox{0.75}{%
\newcommand{\ChrfengXron}[2]{\gradientcell{#1}{41.19194346816732}{45.91583212936492}{Cyan}{Yellow}{60}{#2}}
\begin{tabular}{lll ccc@{\hskip 18pt}c}
\toprule
\multicolumn{3}{l}{\multirow{2}{*}{\Large eng-ron}} & \multicolumn{3}{c}{Vocabulary\hspace*{15pt}} & \multirow{2}{*}{Avg.} \\ \cmidrule(lr{18pt}){4-6}
& & & 4k & 8k & 16k & \\ \midrule

\multirow{4}{*}{\rotatebox[origin=c]{90}{Word-like}} & \multirow{2}{*}{BPE} & Orig. & \ChrfengXron{44.4}{\hphantom{-}} & \ChrfengXron{44.6}{\hphantom{-}} & \ChrfengXron{44.5}{\hphantom{-}} & \ChrfengXron{44.5}{\hphantom{-}} \\
& & Ours & \ChrfengXron{44.7}{\hphantom{-}} & \ChrfengXron{45.3}{\hphantom{-}} & \ChrfengXron{45.1}{\hphantom{-}} & \ChrfengXron{45.1}{\hphantom{-}} \\ \cmidrule{2-7}
& \multirow{2}{*}{Uni.} & Orig. & \ChrfengXron{43.8}{\hphantom{-}} & \ChrfengXron{45.4}{\hphantom{-}} & \ChrfengXron{44.8}{\hphantom{-}} & \ChrfengXron{44.7}{\hphantom{-}} \\
& & Ours & \ChrfengXron{44.3}{\hphantom{-}} & \ChrfengXron{45.1}{\hphantom{-}} & \ChrfengXron{44.5}{\hphantom{-}} & \ChrfengXron{44.6}{\hphantom{-}} \\ \midrule
\multirow{4}{*}{\rotatebox[origin=c]{90}{Morfessor}} & \multirow{2}{*}{BPE} & Orig. & \ChrfengXron{42.7}{\hphantom{-}} & \ChrfengXron{42.5}{\hphantom{-}} & \ChrfengXron{42.4}{\hphantom{-}} & \ChrfengXron{42.5}{\hphantom{-}} \\
& & Ours & \ChrfengXron{42.9}{\hphantom{-}} & \ChrfengXron{44.6}{\hphantom{-}} & \ChrfengXron{44.5}{\hphantom{-}} & \ChrfengXron{44.0}{\hphantom{-}} \\ \cmidrule{2-7}
& \multirow{2}{*}{Uni.} & Orig. & \ChrfengXron{41.7}{\hphantom{-}} & \ChrfengXron{42.4}{\hphantom{-}} & \ChrfengXron{43.0}{\hphantom{-}} & \ChrfengXron{42.4}{\hphantom{-}} \\
& & Ours & \ChrfengXron{43.6}{\hphantom{-}} & \ChrfengXron{44.2}{\hphantom{-}} & \ChrfengXron{43.8}{\hphantom{-}} & \ChrfengXron{43.9}{\hphantom{-}} \\ \bottomrule
\end{tabular}
}\end{minipage}
\\[1.8ex]
\begin{minipage}{.33\textwidth}\scalebox{0.75}{%
\newcommand{\ChrfronXeng}[2]{\gradientcell{#1}{43.76604377522577}{49.01063367214984}{Cyan}{Yellow}{60}{#2}}
\begin{tabular}{lll ccc@{\hskip 18pt}c}
\toprule
\multicolumn{3}{l}{\multirow{2}{*}{\Large ron-eng}} & \multicolumn{3}{c}{Vocabulary\hspace*{15pt}} & \multirow{2}{*}{Avg.} \\ \cmidrule(lr{18pt}){4-6}
& & & 4k & 8k & 16k & \\ \midrule

\multirow{4}{*}{\rotatebox[origin=c]{90}{Word-like}} & \multirow{2}{*}{BPE} & Orig. & \ChrfronXeng{46.5}{\hphantom{-}} & \ChrfronXeng{47.1}{\hphantom{-}} & \ChrfronXeng{47.5}{\hphantom{-}} & \ChrfronXeng{47.0}{\hphantom{-}} \\
& & Ours & \ChrfronXeng{47.5}{\hphantom{-}} & \ChrfronXeng{47.0}{\hphantom{-}} & \ChrfronXeng{48.5}{\hphantom{-}} & \ChrfronXeng{47.7}{\hphantom{-}} \\ \cmidrule{2-7}
& \multirow{2}{*}{Uni.} & Orig. & \ChrfronXeng{47.2}{\hphantom{-}} & \ChrfronXeng{47.5}{\hphantom{-}} & \ChrfronXeng{48.2}{\hphantom{-}} & \ChrfronXeng{47.7}{\hphantom{-}} \\
& & Ours & \ChrfronXeng{46.5}{\hphantom{-}} & \ChrfronXeng{46.6}{\hphantom{-}} & \ChrfronXeng{47.6}{\hphantom{-}} & \ChrfronXeng{46.9}{\hphantom{-}} \\ \midrule
\multirow{4}{*}{\rotatebox[origin=c]{90}{Morfessor}} & \multirow{2}{*}{BPE} & Orig. & \ChrfronXeng{45.4}{\hphantom{-}} & \ChrfronXeng{45.2}{\hphantom{-}} & \ChrfronXeng{45.4}{\hphantom{-}} & \ChrfronXeng{45.3}{\hphantom{-}} \\
& & Ours & \ChrfronXeng{46.3}{\hphantom{-}} & \ChrfronXeng{46.8}{\hphantom{-}} & \ChrfronXeng{47.1}{\hphantom{-}} & \ChrfronXeng{46.7}{\hphantom{-}} \\ \cmidrule{2-7}
& \multirow{2}{*}{Uni.} & Orig. & \ChrfronXeng{44.3}{\hphantom{-}} & \ChrfronXeng{45.5}{\hphantom{-}} & \ChrfronXeng{44.6}{\hphantom{-}} & \ChrfronXeng{44.8}{\hphantom{-}} \\
& & Ours & \ChrfronXeng{46.4}{\hphantom{-}} & \ChrfronXeng{46.0}{\hphantom{-}} & \ChrfronXeng{46.7}{\hphantom{-}} & \ChrfronXeng{46.4}{\hphantom{-}} \\ \bottomrule
\end{tabular}
}\end{minipage}
\begin{minipage}{.33\textwidth}\scalebox{0.75}{%
\newcommand{\ChrfengXita}[2]{\gradientcell{#1}{44.8508240120588}{48.73276123793681}{Cyan}{Yellow}{60}{#2}}
\begin{tabular}{lll ccc@{\hskip 18pt}c}
\toprule
\multicolumn{3}{l}{\multirow{2}{*}{\Large eng-ita}} & \multicolumn{3}{c}{Vocabulary\hspace*{15pt}} & \multirow{2}{*}{Avg.} \\ \cmidrule(lr{18pt}){4-6}
& & & 4k & 8k & 16k & \\ \midrule

\multirow{4}{*}{\rotatebox[origin=c]{90}{Word-like}} & \multirow{2}{*}{BPE} & Orig. & \ChrfengXita{46.7}{\hphantom{-}} & \ChrfengXita{46.8}{\hphantom{-}} & \ChrfengXita{47.7}{\hphantom{-}} & \ChrfengXita{47.1}{\hphantom{-}} \\
& & Ours & \ChrfengXita{46.1}{\hphantom{-}} & \ChrfengXita{47.1}{\hphantom{-}} & \ChrfengXita{47.1}{\hphantom{-}} & \ChrfengXita{46.8}{\hphantom{-}} \\ \cmidrule{2-7}
& \multirow{2}{*}{Uni.} & Orig. & \ChrfengXita{46.7}{\hphantom{-}} & \ChrfengXita{48.2}{\hphantom{-}} & \ChrfengXita{47.8}{\hphantom{-}} & \ChrfengXita{47.6}{\hphantom{-}} \\
& & Ours & \ChrfengXita{46.8}{\hphantom{-}} & \ChrfengXita{47.8}{\hphantom{-}} & \ChrfengXita{47.5}{\hphantom{-}} & \ChrfengXita{47.3}{\hphantom{-}} \\ \midrule
\multirow{4}{*}{\rotatebox[origin=c]{90}{Morfessor}} & \multirow{2}{*}{BPE} & Orig. & \ChrfengXita{46.2}{\hphantom{-}} & \ChrfengXita{45.8}{\hphantom{-}} & \ChrfengXita{45.6}{\hphantom{-}} & \ChrfengXita{45.9}{\hphantom{-}} \\
& & Ours & \ChrfengXita{46.5}{\hphantom{-}} & \ChrfengXita{47.2}{\hphantom{-}} & \ChrfengXita{47.7}{\hphantom{-}} & \ChrfengXita{47.2}{\hphantom{-}} \\ \cmidrule{2-7}
& \multirow{2}{*}{Uni.} & Orig. & \ChrfengXita{45.7}{\hphantom{-}} & \ChrfengXita{46.1}{\hphantom{-}} & \ChrfengXita{45.7}{\hphantom{-}} & \ChrfengXita{45.8}{\hphantom{-}} \\
& & Ours & \ChrfengXita{46.0}{\hphantom{-}} & \ChrfengXita{47.3}{\hphantom{-}} & \ChrfengXita{45.4}{\hphantom{-}} & \ChrfengXita{46.2}{\hphantom{-}} \\ \bottomrule
\end{tabular}
}\end{minipage}
\begin{minipage}{.33\textwidth}\scalebox{0.75}{%
\newcommand{\ChrfitaXeng}[2]{\gradientcell{#1}{44.658553145579724}{48.73641457522783}{Cyan}{Yellow}{60}{#2}}
\begin{tabular}{lll ccc@{\hskip 18pt}c}
\toprule
\multicolumn{3}{l}{\multirow{2}{*}{\Large ita-eng}} & \multicolumn{3}{c}{Vocabulary\hspace*{15pt}} & \multirow{2}{*}{Avg.} \\ \cmidrule(lr{18pt}){4-6}
& & & 4k & 8k & 16k & \\ \midrule

\multirow{4}{*}{\rotatebox[origin=c]{90}{Word-like}} & \multirow{2}{*}{BPE} & Orig. & \ChrfitaXeng{46.6}{\hphantom{-}} & \ChrfitaXeng{47.3}{\hphantom{-}} & \ChrfitaXeng{48.2}{\hphantom{-}} & \ChrfitaXeng{47.4}{\hphantom{-}} \\
& & Ours & \ChrfitaXeng{46.7}{\hphantom{-}} & \ChrfitaXeng{47.5}{\hphantom{-}} & \ChrfitaXeng{47.4}{\hphantom{-}} & \ChrfitaXeng{47.2}{\hphantom{-}} \\ \cmidrule{2-7}
& \multirow{2}{*}{Uni.} & Orig. & \ChrfitaXeng{47.5}{\hphantom{-}} & \ChrfitaXeng{47.6}{\hphantom{-}} & \ChrfitaXeng{47.7}{\hphantom{-}} & \ChrfitaXeng{47.6}{\hphantom{-}} \\
& & Ours & \ChrfitaXeng{46.5}{\hphantom{-}} & \ChrfitaXeng{47.7}{\hphantom{-}} & \ChrfitaXeng{47.6}{\hphantom{-}} & \ChrfitaXeng{47.3}{\hphantom{-}} \\ \midrule
\multirow{4}{*}{\rotatebox[origin=c]{90}{Morfessor}} & \multirow{2}{*}{BPE} & Orig. & \ChrfitaXeng{45.4}{\hphantom{-}} & \ChrfitaXeng{46.0}{\hphantom{-}} & \ChrfitaXeng{45.2}{\hphantom{-}} & \ChrfitaXeng{45.5}{\hphantom{-}} \\
& & Ours & \ChrfitaXeng{46.2}{\hphantom{-}} & \ChrfitaXeng{46.5}{\hphantom{-}} & \ChrfitaXeng{46.9}{\hphantom{-}} & \ChrfitaXeng{46.5}{\hphantom{-}} \\ \cmidrule{2-7}
& \multirow{2}{*}{Uni.} & Orig. & \ChrfitaXeng{45.2}{\hphantom{-}} & \ChrfitaXeng{45.3}{\hphantom{-}} & \ChrfitaXeng{45.4}{\hphantom{-}} & \ChrfitaXeng{45.3}{\hphantom{-}} \\
& & Ours & \ChrfitaXeng{46.5}{\hphantom{-}} & \ChrfitaXeng{46.7}{\hphantom{-}} & \ChrfitaXeng{46.2}{\hphantom{-}} & \ChrfitaXeng{46.5}{\hphantom{-}} \\ \bottomrule
\end{tabular}
}\end{minipage}
\\[1.8ex]
\begin{minipage}{.33\textwidth}\scalebox{0.75}{%
\newcommand{\ChrfitaXnld}[2]{\gradientcell{#1}{33.84541769351303}{37.6651628963067}{Cyan}{Yellow}{60}{#2}}
\begin{tabular}{lll ccc@{\hskip 18pt}c}
\toprule
\multicolumn{3}{l}{\multirow{2}{*}{\Large ita-nld}} & \multicolumn{3}{c}{Vocabulary\hspace*{15pt}} & \multirow{2}{*}{Avg.} \\ \cmidrule(lr{18pt}){4-6}
& & & 4k & 8k & 16k & \\ \midrule

\multirow{4}{*}{\rotatebox[origin=c]{90}{Word-like}} & \multirow{2}{*}{BPE} & Orig. & \ChrfitaXnld{36.3}{\hphantom{-}} & \ChrfitaXnld{35.8}{\hphantom{-}} & \ChrfitaXnld{36.3}{\hphantom{-}} & \ChrfitaXnld{36.1}{\hphantom{-}} \\
& & Ours & \ChrfitaXnld{35.9}{\hphantom{-}} & \ChrfitaXnld{36.1}{\hphantom{-}} & \ChrfitaXnld{37.2}{\hphantom{-}} & \ChrfitaXnld{36.4}{\hphantom{-}} \\ \cmidrule{2-7}
& \multirow{2}{*}{Uni.} & Orig. & \ChrfitaXnld{35.9}{\hphantom{-}} & \ChrfitaXnld{36.2}{\hphantom{-}} & \ChrfitaXnld{37.0}{\hphantom{-}} & \ChrfitaXnld{36.3}{\hphantom{-}} \\
& & Ours & \ChrfitaXnld{35.3}{\hphantom{-}} & \ChrfitaXnld{35.9}{\hphantom{-}} & \ChrfitaXnld{36.9}{\hphantom{-}} & \ChrfitaXnld{36.0}{\hphantom{-}} \\ \midrule
\multirow{4}{*}{\rotatebox[origin=c]{90}{Morfessor}} & \multirow{2}{*}{BPE} & Orig. & \ChrfitaXnld{35.4}{\hphantom{-}} & \ChrfitaXnld{35.4}{\hphantom{-}} & \ChrfitaXnld{34.3}{\hphantom{-}} & \ChrfitaXnld{35.1}{\hphantom{-}} \\
& & Ours & \ChrfitaXnld{34.7}{\hphantom{-}} & \ChrfitaXnld{36.3}{\hphantom{-}} & \ChrfitaXnld{36.6}{\hphantom{-}} & \ChrfitaXnld{35.9}{\hphantom{-}} \\ \cmidrule{2-7}
& \multirow{2}{*}{Uni.} & Orig. & \ChrfitaXnld{35.0}{\hphantom{-}} & \ChrfitaXnld{35.2}{\hphantom{-}} & \ChrfitaXnld{35.6}{\hphantom{-}} & \ChrfitaXnld{35.3}{\hphantom{-}} \\
& & Ours & \ChrfitaXnld{36.8}{\hphantom{-}} & \ChrfitaXnld{36.5}{\hphantom{-}} & \ChrfitaXnld{36.4}{\hphantom{-}} & \ChrfitaXnld{36.6}{\hphantom{-}} \\ \bottomrule
\end{tabular}
}\end{minipage}
\begin{minipage}{.33\textwidth}\scalebox{0.75}{%
\newcommand{\ChrfnldXita}[2]{\gradientcell{#1}{34.82488787489153}{38.416866363599524}{Cyan}{Yellow}{60}{#2}}
\begin{tabular}{lll ccc@{\hskip 18pt}c}
\toprule
\multicolumn{3}{l}{\multirow{2}{*}{\Large nld-ita}} & \multicolumn{3}{c}{Vocabulary\hspace*{15pt}} & \multirow{2}{*}{Avg.} \\ \cmidrule(lr{18pt}){4-6}
& & & 4k & 8k & 16k & \\ \midrule

\multirow{4}{*}{\rotatebox[origin=c]{90}{Word-like}} & \multirow{2}{*}{BPE} & Orig. & \ChrfnldXita{36.5}{\hphantom{-}} & \ChrfnldXita{37.0}{\hphantom{-}} & \ChrfnldXita{37.1}{\hphantom{-}} & \ChrfnldXita{36.9}{\hphantom{-}} \\
& & Ours & \ChrfnldXita{36.3}{\hphantom{-}} & \ChrfnldXita{36.7}{\hphantom{-}} & \ChrfnldXita{37.8}{\hphantom{-}} & \ChrfnldXita{36.9}{\hphantom{-}} \\ \cmidrule{2-7}
& \multirow{2}{*}{Uni.} & Orig. & \ChrfnldXita{36.3}{\hphantom{-}} & \ChrfnldXita{37.5}{\hphantom{-}} & \ChrfnldXita{37.2}{\hphantom{-}} & \ChrfnldXita{37.0}{\hphantom{-}} \\
& & Ours & \ChrfnldXita{36.8}{\hphantom{-}} & \ChrfnldXita{35.5}{\hphantom{-}} & \ChrfnldXita{37.3}{\hphantom{-}} & \ChrfnldXita{36.6}{\hphantom{-}} \\ \midrule
\multirow{4}{*}{\rotatebox[origin=c]{90}{Morfessor}} & \multirow{2}{*}{BPE} & Orig. & \ChrfnldXita{35.3}{\hphantom{-}} & \ChrfnldXita{36.1}{\hphantom{-}} & \ChrfnldXita{36.1}{\hphantom{-}} & \ChrfnldXita{35.8}{\hphantom{-}} \\
& & Ours & \ChrfnldXita{37.4}{\hphantom{-}} & \ChrfnldXita{36.8}{\hphantom{-}} & \ChrfnldXita{37.9}{\hphantom{-}} & \ChrfnldXita{37.4}{\hphantom{-}} \\ \cmidrule{2-7}
& \multirow{2}{*}{Uni.} & Orig. & \ChrfnldXita{35.7}{\hphantom{-}} & \ChrfnldXita{35.7}{\hphantom{-}} & \ChrfnldXita{36.3}{\hphantom{-}} & \ChrfnldXita{35.9}{\hphantom{-}} \\
& & Ours & \ChrfnldXita{36.7}{\hphantom{-}} & \ChrfnldXita{36.8}{\hphantom{-}} & \ChrfnldXita{36.2}{\hphantom{-}} & \ChrfnldXita{36.6}{\hphantom{-}} \\ \bottomrule
\end{tabular}
}\end{minipage}
\begin{minipage}{.33\textwidth}\scalebox{0.75}{%
\newcommand{\ChrfronXita}[2]{\gradientcell{#1}{38.05187137032875}{41.5731608121849}{Cyan}{Yellow}{60}{#2}}
\begin{tabular}{lll ccc@{\hskip 18pt}c}
\toprule
\multicolumn{3}{l}{\multirow{2}{*}{\Large ron-ita}} & \multicolumn{3}{c}{Vocabulary\hspace*{15pt}} & \multirow{2}{*}{Avg.} \\ \cmidrule(lr{18pt}){4-6}
& & & 4k & 8k & 16k & \\ \midrule

\multirow{4}{*}{\rotatebox[origin=c]{90}{Word-like}} & \multirow{2}{*}{BPE} & Orig. & \ChrfronXita{40.2}{\hphantom{-}} & \ChrfronXita{41.1}{\hphantom{-}} & \ChrfronXita{40.7}{\hphantom{-}} & \ChrfronXita{40.7}{\hphantom{-}} \\
& & Ours & \ChrfronXita{41.0}{\hphantom{-}} & \ChrfronXita{40.8}{\hphantom{-}} & \ChrfronXita{40.4}{\hphantom{-}} & \ChrfronXita{40.8}{\hphantom{-}} \\ \cmidrule{2-7}
& \multirow{2}{*}{Uni.} & Orig. & \ChrfronXita{40.0}{\hphantom{-}} & \ChrfronXita{40.1}{\hphantom{-}} & \ChrfronXita{40.5}{\hphantom{-}} & \ChrfronXita{40.2}{\hphantom{-}} \\
& & Ours & \ChrfronXita{39.8}{\hphantom{-}} & \ChrfronXita{40.9}{\hphantom{-}} & \ChrfronXita{39.8}{\hphantom{-}} & \ChrfronXita{40.2}{\hphantom{-}} \\ \midrule
\multirow{4}{*}{\rotatebox[origin=c]{90}{Morfessor}} & \multirow{2}{*}{BPE} & Orig. & \ChrfronXita{39.4}{\hphantom{-}} & \ChrfronXita{38.8}{\hphantom{-}} & \ChrfronXita{38.6}{\hphantom{-}} & \ChrfronXita{38.9}{\hphantom{-}} \\
& & Ours & \ChrfronXita{40.3}{\hphantom{-}} & \ChrfronXita{40.3}{\hphantom{-}} & \ChrfronXita{40.4}{\hphantom{-}} & \ChrfronXita{40.3}{\hphantom{-}} \\ \cmidrule{2-7}
& \multirow{2}{*}{Uni.} & Orig. & \ChrfronXita{39.0}{\hphantom{-}} & \ChrfronXita{39.4}{\hphantom{-}} & \ChrfronXita{39.1}{\hphantom{-}} & \ChrfronXita{39.2}{\hphantom{-}} \\
& & Ours & \ChrfronXita{40.8}{\hphantom{-}} & \ChrfronXita{40.0}{\hphantom{-}} & \ChrfronXita{40.3}{\hphantom{-}} & \ChrfronXita{40.4}{\hphantom{-}} \\ \bottomrule
\end{tabular}
}\end{minipage}
\\[1.8ex]
\begin{minipage}{.33\textwidth}\scalebox{0.75}{%
\newcommand{\ChrfitaXron}[2]{\gradientcell{#1}{34.46546238157449}{38.794302090051055}{Cyan}{Yellow}{60}{#2}}
\begin{tabular}{lll ccc@{\hskip 18pt}c}
\toprule
\multicolumn{3}{l}{\multirow{2}{*}{\Large ita-ron}} & \multicolumn{3}{c}{Vocabulary\hspace*{15pt}} & \multirow{2}{*}{Avg.} \\ \cmidrule(lr{18pt}){4-6}
& & & 4k & 8k & 16k & \\ \midrule

\multirow{4}{*}{\rotatebox[origin=c]{90}{Word-like}} & \multirow{2}{*}{BPE} & Orig. & \ChrfitaXron{37.5}{\hphantom{-}} & \ChrfitaXron{37.9}{\hphantom{-}} & \ChrfitaXron{38.3}{\hphantom{-}} & \ChrfitaXron{37.9}{\hphantom{-}} \\
& & Ours & \ChrfitaXron{37.8}{\hphantom{-}} & \ChrfitaXron{38.1}{\hphantom{-}} & \ChrfitaXron{38.0}{\hphantom{-}} & \ChrfitaXron{37.9}{\hphantom{-}} \\ \cmidrule{2-7}
& \multirow{2}{*}{Uni.} & Orig. & \ChrfitaXron{37.0}{\hphantom{-}} & \ChrfitaXron{38.1}{\hphantom{-}} & \ChrfitaXron{37.7}{\hphantom{-}} & \ChrfitaXron{37.6}{\hphantom{-}} \\
& & Ours & \ChrfitaXron{37.6}{\hphantom{-}} & \ChrfitaXron{37.7}{\hphantom{-}} & \ChrfitaXron{37.6}{\hphantom{-}} & \ChrfitaXron{37.6}{\hphantom{-}} \\ \midrule
\multirow{4}{*}{\rotatebox[origin=c]{90}{Morfessor}} & \multirow{2}{*}{BPE} & Orig. & \ChrfitaXron{35.6}{\hphantom{-}} & \ChrfitaXron{36.0}{\hphantom{-}} & \ChrfitaXron{36.3}{\hphantom{-}} & \ChrfitaXron{36.0}{\hphantom{-}} \\
& & Ours & \ChrfitaXron{38.2}{\hphantom{-}} & \ChrfitaXron{38.2}{\hphantom{-}} & \ChrfitaXron{37.8}{\hphantom{-}} & \ChrfitaXron{38.1}{\hphantom{-}} \\ \cmidrule{2-7}
& \multirow{2}{*}{Uni.} & Orig. & \ChrfitaXron{35.0}{\hphantom{-}} & \ChrfitaXron{35.5}{\hphantom{-}} & \ChrfitaXron{35.5}{\hphantom{-}} & \ChrfitaXron{35.3}{\hphantom{-}} \\
& & Ours & \ChrfitaXron{37.4}{\hphantom{-}} & \ChrfitaXron{37.3}{\hphantom{-}} & \ChrfitaXron{37.0}{\hphantom{-}} & \ChrfitaXron{37.2}{\hphantom{-}} \\ \bottomrule
\end{tabular}
}\end{minipage}
\begin{minipage}{.33\textwidth}\scalebox{0.75}{%
\newcommand{\ChrfronXnld}[2]{\gradientcell{#1}{34.114526717393005}{37.549137854053626}{Cyan}{Yellow}{60}{#2}}
\begin{tabular}{lll ccc@{\hskip 18pt}c}
\toprule
\multicolumn{3}{l}{\multirow{2}{*}{\Large ron-nld}} & \multicolumn{3}{c}{Vocabulary\hspace*{15pt}} & \multirow{2}{*}{Avg.} \\ \cmidrule(lr{18pt}){4-6}
& & & 4k & 8k & 16k & \\ \midrule

\multirow{4}{*}{\rotatebox[origin=c]{90}{Word-like}} & \multirow{2}{*}{BPE} & Orig. & \ChrfronXnld{36.0}{\hphantom{-}} & \ChrfronXnld{35.9}{\hphantom{-}} & \ChrfronXnld{37.0}{\hphantom{-}} & \ChrfronXnld{36.3}{\hphantom{-}} \\
& & Ours & \ChrfronXnld{35.7}{\hphantom{-}} & \ChrfronXnld{36.5}{\hphantom{-}} & \ChrfronXnld{36.7}{\hphantom{-}} & \ChrfronXnld{36.3}{\hphantom{-}} \\ \cmidrule{2-7}
& \multirow{2}{*}{Uni.} & Orig. & \ChrfronXnld{35.7}{\hphantom{-}} & \ChrfronXnld{36.0}{\hphantom{-}} & \ChrfronXnld{36.5}{\hphantom{-}} & \ChrfronXnld{36.1}{\hphantom{-}} \\
& & Ours & \ChrfronXnld{35.1}{\hphantom{-}} & \ChrfronXnld{35.2}{\hphantom{-}} & \ChrfronXnld{36.6}{\hphantom{-}} & \ChrfronXnld{35.6}{\hphantom{-}} \\ \midrule
\multirow{4}{*}{\rotatebox[origin=c]{90}{Morfessor}} & \multirow{2}{*}{BPE} & Orig. & \ChrfronXnld{34.7}{\hphantom{-}} & \ChrfronXnld{35.9}{\hphantom{-}} & \ChrfronXnld{35.2}{\hphantom{-}} & \ChrfronXnld{35.3}{\hphantom{-}} \\
& & Ours & \ChrfronXnld{36.1}{\hphantom{-}} & \ChrfronXnld{36.0}{\hphantom{-}} & \ChrfronXnld{36.1}{\hphantom{-}} & \ChrfronXnld{36.1}{\hphantom{-}} \\ \cmidrule{2-7}
& \multirow{2}{*}{Uni.} & Orig. & \ChrfronXnld{35.3}{\hphantom{-}} & \ChrfronXnld{34.6}{\hphantom{-}} & \ChrfronXnld{35.1}{\hphantom{-}} & \ChrfronXnld{35.0}{\hphantom{-}} \\
& & Ours & \ChrfronXnld{35.6}{\hphantom{-}} & \ChrfronXnld{36.8}{\hphantom{-}} & \ChrfronXnld{35.6}{\hphantom{-}} & \ChrfronXnld{36.0}{\hphantom{-}} \\ \bottomrule
\end{tabular}
}\end{minipage}
\begin{minipage}{.33\textwidth}\scalebox{0.75}{%
\newcommand{\ChrfnldXron}[2]{\gradientcell{#1}{31.805987165733782}{35.67235132648232}{Cyan}{Yellow}{60}{#2}}
\begin{tabular}{lll ccc@{\hskip 18pt}c}
\toprule
\multicolumn{3}{l}{\multirow{2}{*}{\Large nld-ron}} & \multicolumn{3}{c}{Vocabulary\hspace*{15pt}} & \multirow{2}{*}{Avg.} \\ \cmidrule(lr{18pt}){4-6}
& & & 4k & 8k & 16k & \\ \midrule

\multirow{4}{*}{\rotatebox[origin=c]{90}{Word-like}} & \multirow{2}{*}{BPE} & Orig. & \ChrfnldXron{33.6}{\hphantom{-}} & \ChrfnldXron{33.5}{\hphantom{-}} & \ChrfnldXron{34.7}{\hphantom{-}} & \ChrfnldXron{33.9}{\hphantom{-}} \\
& & Ours & \ChrfnldXron{33.6}{\hphantom{-}} & \ChrfnldXron{34.1}{\hphantom{-}} & \ChrfnldXron{35.1}{\hphantom{-}} & \ChrfnldXron{34.3}{\hphantom{-}} \\ \cmidrule{2-7}
& \multirow{2}{*}{Uni.} & Orig. & \ChrfnldXron{33.4}{\hphantom{-}} & \ChrfnldXron{34.7}{\hphantom{-}} & \ChrfnldXron{35.0}{\hphantom{-}} & \ChrfnldXron{34.4}{\hphantom{-}} \\
& & Ours & \ChrfnldXron{33.5}{\hphantom{-}} & \ChrfnldXron{35.2}{\hphantom{-}} & \ChrfnldXron{34.4}{\hphantom{-}} & \ChrfnldXron{34.3}{\hphantom{-}} \\ \midrule
\multirow{4}{*}{\rotatebox[origin=c]{90}{Morfessor}} & \multirow{2}{*}{BPE} & Orig. & \ChrfnldXron{32.7}{\hphantom{-}} & \ChrfnldXron{32.9}{\hphantom{-}} & \ChrfnldXron{33.1}{\hphantom{-}} & \ChrfnldXron{32.9}{\hphantom{-}} \\
& & Ours & \ChrfnldXron{33.6}{\hphantom{-}} & \ChrfnldXron{34.8}{\hphantom{-}} & \ChrfnldXron{33.6}{\hphantom{-}} & \ChrfnldXron{34.0}{\hphantom{-}} \\ \cmidrule{2-7}
& \multirow{2}{*}{Uni.} & Orig. & \ChrfnldXron{32.3}{\hphantom{-}} & \ChrfnldXron{32.8}{\hphantom{-}} & \ChrfnldXron{32.3}{\hphantom{-}} & \ChrfnldXron{32.5}{\hphantom{-}} \\
& & Ours & \ChrfnldXron{33.1}{\hphantom{-}} & \ChrfnldXron{33.9}{\hphantom{-}} & \ChrfnldXron{33.7}{\hphantom{-}} & \ChrfnldXron{33.6}{\hphantom{-}} \\ \bottomrule
\end{tabular}
}\end{minipage}
\\[1.8ex]

\caption{The chrF scores for 18 language pairs of the IWSLT 2017. The blue-yellow scale is fit to the value range across each table.}\label{tab:chrf_all}

\end{table*}
\end{document}